\documentclass[10pt,twocolumn,letterpaper]{article}

\usepackage[accsupp]{axessibility}  %

\usepackage[pagenumbers]{cvpr} %

\definecolor{cvprblue}{rgb}{0.21,0.49,0.74}
\usepackage{multirow,makecell}
\usepackage[table]{xcolor}
\usepackage[pagebackref,breaklinks,colorlinks,allcolors=cvprblue]{hyperref}

\usepackage{arydshln} %
\usepackage{dblfloatfix} %
\usepackage{placeins}    %
\def\paperID{*****} %
\def\confName{CVPR}
\def\confYear{2026}

\title{Depth Adaptive Efficient Visual Autoregressive Modeling}

\author{
Chunliang Li$^{1}$\thanks{Equal contribution} \quad 
Tianze Cao$^{1}$\footnotemark[1] \quad
Sanyuan Zhao$^{1,2}$\thanks{Corresponding author}\\
$^{1}$Beijing Institute of Technology\quad $^{2}$Beijing Institute of Technology, Zhuhai\\
{\tt\small jbjic@icloud.com \quad \{caotianze, zhaosanyuan\}@bit.edu.cn}
}

\begin{document}
\maketitle
\begin{abstract}

Visual Autoregressive (VAR) modeling inefficiently applies a fixed computational depth to each position when generating high-resolution images. While existing methods accelerate inference by pruning tokens using frequency maps, their binary hard-pruning approach is fundamentally limited and fails to improve quality even with better frequency estimation. 
Observing that VAR models possess significant depth redundancy, we propose a paradigm shift from pruning entire tokens to adaptively allocating per-token computational depth. To this end, we introduce DepthVAR, a training-free framework that dynamically allocates computation.
It integrates an adaptive depth scheduler, which assigns computational depth via a cyclic rotated schedule for balanced, non-static refinement, with a dynamic inference process that translates these depths into layer-major masks, selectively applies transformer blocks, and blends the resulting codes to ensure each token's influence is proportional to its processing depth.
Extensive experiments show that DepthVAR achieves 2.3$\times$-3.1$\times$ acceleration with minimal quality loss, offering a competitive compute-performance trade-off compared to existing hard-pruning approaches. 
Code is available at \href{https://github.com/STOVAGtz/DepthVAR}{https://github.com/STOVAGtz/DepthVAR}.

\end{abstract}

\section{Introduction}

Recent advances in Autoregressive (AR) models use `next-token' prediction to enable the step-by-step synthesis of complex visual structures, offering a unified probabilistic framework for text-to-image generation. However, the sequential nature of AR modeling leads to long token chains, which increases computational cost and memory demands, especially for high-resolution images. Visual Autoregressive (VAR) modeling~\cite{tianVisualAutoregressiveModeling,hanInfinityScalingBitwise2025,tangHARTEfficientVisual2024a} mitigates this issue by shifting to `next-scale' prediction and generating images hierarchically across scales. This substantially reduces sequence length and prediction latency, yet tokens grow quadratically with each scale, creating significant computational overhead from the inefficient uniform processing of tokens that represent regions requiring less computation.

\begin{figure}[t]
  \centering
  \includegraphics[width=\linewidth]{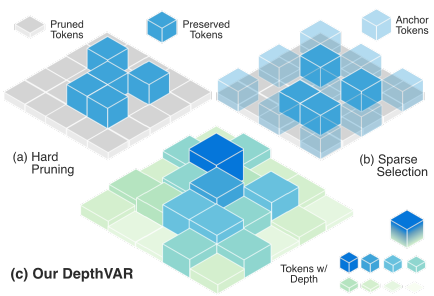}
    \caption{\textbf{Comparison of VAR acceleration paradigms.} (a) Hard token pruning (e.g.~\cite{guoFastVARLinearVisual2025}) discards tokens. (b) Sparse token selection (e.g.~\cite{chenFrequencyAwareAutoregressiveModeling2025}) retains anchor tokens to preserve background structure. (c) We adaptively vary the layers processed per token.}
    \label{fig:paradigm}
    \vspace{-1.5em}
\end{figure}

\begin{figure*}[ht]
  \centering
  \begin{subfigure}[t]{0.34\textwidth}
    \includegraphics[width=\linewidth, trim=10 10 10 10, clip]{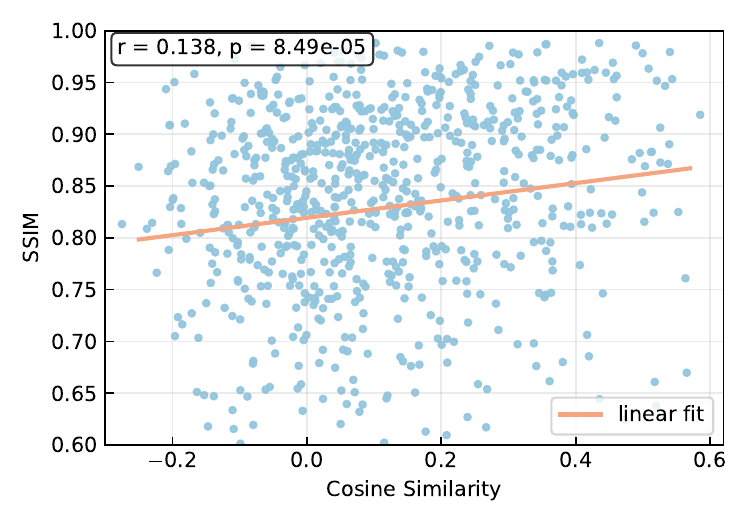}
    \caption{Frequency approximation vs. image quality.}
    \label{fig:freq_approx}
  \end{subfigure}
  \begin{subfigure}[t]{0.65\textwidth}
    \includegraphics[width=\linewidth]{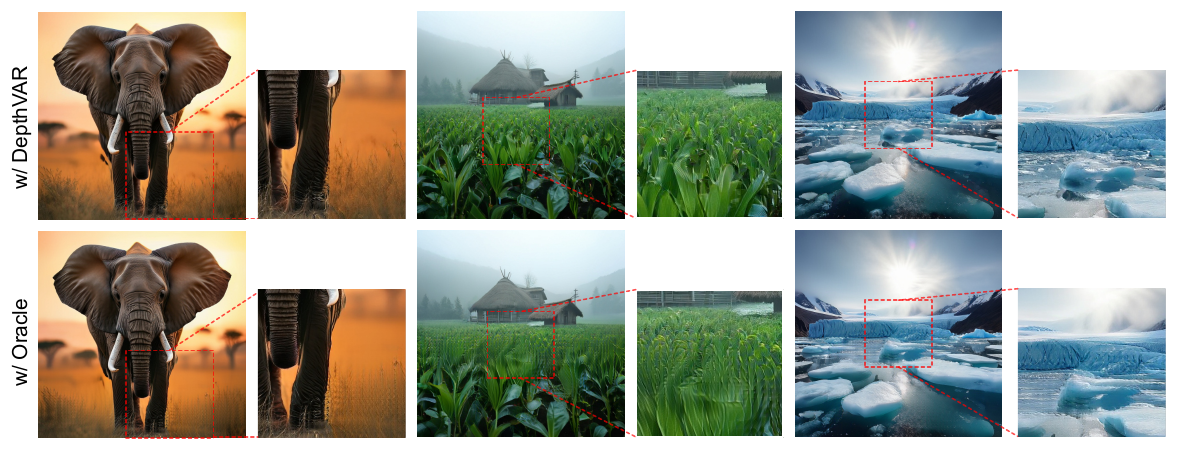}
    \caption{Qualitative results of hard-pruning with an oracle frequency mask.}
    \label{fig:freq_oracle}
  \end{subfigure}
      \caption{\textbf{Limitations of frequency-based pruning.} (a) The accuracy of frequency map approximation, a common heuristic for token pruning, correlates poorly with final image quality.
      (b) Employing a perfect oracle frequency mask for hard-pruning still results in significant quality degradation, which points to an inherent flaw in the strategy.}
  \label{fig:freq_limit}
  \vspace{-1em}
\end{figure*}

In pursuit of non-uniform processing for efficient VAR inference, prior works~\cite{guoFastVARLinearVisual2025, chenFrequencyAwareAutoregressiveModeling2025} prune tokens at larger scales, assuming high-frequency tokens are more critical for later-stage refinement. 
These methods estimate high-frequency distributions from intermediate outputs to identify and discard less important tokens. However, such hand-crafted frequency estimations are often inaccurate, degrading quality in pruned regions, which is sometimes mitigated with sparse background grids~\cite{chenFrequencyAwareAutoregressiveModeling2025}.
Specifically, as shown in \cref{fig:freq_limit}, we find that more accurate frequency estimation does not guarantee improved generation quality, suggesting a fundamental limitation in the hard-pruning paradigm.%

These frequency-based methods~\cite{guoFastVARLinearVisual2025, chenFrequencyAwareAutoregressiveModeling2025} dichotomize tokens into a `keep' or `prune' status.
This motivates our investigation into a more continuous form of computational scaling, where we find that VAR models exhibit exploitable token-wise and layer-wise depth redundancy, diverging from previous findings~\cite{lifreqexit}, which suggests that reducing per-token depth rather than hard-pruning tokens can potentially save computation while better preserving image quality. 
While depth redundancy~\cite{del2023skipdecode, elhoushi2024layerskip, fanReducingTransformerDepth2019a} motivates modern early exiting~\cite{schusterConfidentAdaptiveLanguage2022, fanReducingTransformerDepth2019a} and dynamic depth~\cite{raposoMixtureofDepthsDynamicallyAllocating2024, he2025router, del2023skipdecode} methods, differences in the saturation behavior make direct transfer to VAR models non-trivial. Accordingly, we hypothesize that computational resources should be allocated in a more continuous manner across tokens, allowing for a non-binary approach to non-uniform processing at larger scales.

Based on these observations, we propose DepthVAR, a training-free inference acceleration framework that precisely controls per-token inference depth. %
As illustrated in \cref{fig:paradigm}, DepthVAR extends beyond prior hard token pruning and sparse selection paradigms by introducing a continuous depth allocation scheme.
To exploit depth redundancy, our dynamic inference framework converts token depth scores into a layer-major mask via bit-reversal to ensure unbiased layer utilization, selectively applies transformer blocks while reusing cached layer behaviors to maintain continuity, and blends the resulting codes to ensure each position's influence is proportional to its processing depth.
These per-position depths are determined by an Adaptive Depth Score Scheduler, which applies a cyclic rotated schedule to prior decision rank maps. This process generates non-static depth scores to ensure balanced, continuous refinement across the image. 
Integrating these components, DepthVAR adaptively achieves efficient and fine-grained dynamic inference. 
We demonstrate that our DepthVAR can accelerate VAR inference with a better performance trade-off than previous hard prune approaches and achieves $2.3\times$-$3.1\times$ acceleration with minimal quality loss.
In a nutshell, our contributions are as follows,

\begin{itemize}
    \item We reveal the limitations of frequency-based hard-pruning and identify exploitable depth redundancy in VAR models as a more effective path toward acceleration.
    \item We propose DepthVAR, a training-free framework that enables continuous depth allocation by integrating a cyclic adaptive depth scheduler with a dynamic inference mechanism for selective computation and code blending.
    \item Experiments on multiple benchmarks demonstrate that DepthVAR achieves $2.3\times$-$3.1\times$ acceleration and marginally superior quality compared with prior hard-pruning acceleration methods.
\end{itemize}

\section{Related Work}

\noindent \textbf{Dynamic Depth in Transformers.} A key strategy for accelerating Transformer inference~\cite{vaswaniAttentionAllYou2017, chitty2023survey, tang2024surveytransformercompression} is to dynamically activate input-specific sub-networks, rather than static methods like quantization~\cite{shen2020q, sun2020mobilebert, bai2021binarybert, liuspinquant} or distillation~\cite{hinton2015distilling, sanh2019distilbert, gu2023minillm}, including Mixture-of-Experts (MoE)~\cite{shazeer2017outrageously, lepikhin2020gshard, fedus2022switch} and adaptive depth methods~\cite{dehghaniUniversalTransformers2019, elbayad2020depth, fanReducingTransformerDepth2019a, schusterConfidentAdaptiveLanguage2022} that vary layer usage per token. 
As a form of adaptive depth, early exiting was first introduced for DNNs and CNNs~\cite{bolukbasi2017adaptive, huang2017multi, teerapittayanon2016branchynet}, and was later applied to encoder-only Transformers like BERT~\cite{devlin2019bert, zhouBERTLosesPatience2020, hou2020dynabert, zhu2021leebert, xin2021berxit, schwartz2020right}, where most methods rely on model confidence scores or small exit classifiers~\cite{schuster2021consistent, xin2021berxit, schwartz2020right}, and subsequently extended to decoder-based LMs~\cite{schusterConfidentAdaptiveLanguage2022, raposoMixtureofDepthsDynamicallyAllocating2024,del2023skipdecode, elhoushi2024layerskip, shan2024early}. 
Specifically, Universal Transformer~\cite{dehghaniUniversalTransformers2019} introduced position-wise stopping with ACT~\cite{gravesAdaptiveComputationTime2017}, while Depth-Adaptive Transformer~\cite{elbayad2020depth} trained auxiliary exits. Furthermore, LayerDrop~\cite{fanReducingTransformerDepth2019a} demonstrated that dropping layers during training enables inference over sub-networks, and CALM~\cite{schusterConfidentAdaptiveLanguage2022} aligns token-level exit decisions with sequence quality targets. 
More flexible approaches learn to allocate depth per token~\cite{raposoMixtureofDepthsDynamicallyAllocating2024, he2025router, del2023skipdecode, fan2025position}, e.g., Mixture-of-Depths(MoD)~\cite{raposoMixtureofDepthsDynamicallyAllocating2024} trained block routers to bypass certain blocks, later extended by Router-Tuning~\cite{he2025router} to a retro-fit framework. Methods like SkipDecode~\cite{del2023skipdecode} and Depth Decay Decoding~\cite{fan2025position} determine depths via decay rules, offering more flexibility over early-exiting. The emergence of similar strategies in vision models~\cite{fei2022deecap, tang2023you, lifreqexit} further underscores their potential for accelerating VARs.

\noindent \textbf{Efficient Visual Autoregressive Modeling.}
The scale-by-scale generation paradigm of Visual Autoregressive Modeling (VAR)~\cite{tianVisualAutoregressiveModeling, hanInfinityScalingBitwise2025, tangHARTEfficientVisual2024a, chenTTSVARTestTimeScaling2025} prevents the direct application of parallel decoding strategies from sequential generation, such as Speculative Decoding~\cite{leviathan2023fast} and Block-wise Parallel Decoding~\cite{stern2018blockwise}.
Theoretically, VAR improves time complexity from $O(n^6)$ to $O(n^4)$~\cite{tianVisualAutoregressiveModeling}, with a potential near-quadratic limit of $O(n^{2+o(1)})$~\cite{keComputationalLimitsProvably2025}, and LiteVAR~\cite{xieLiteVARCompressingVisual2024} achieves near $O(n^2)$ in practice. Some approaches reduce memory~\cite{liMemoryEfficientVisualAutoregressive2025} or employ linear complexity mechanisms like Mamba~\cite{gu2024mamba} to decouple inter-scale computations~\cite{renMVARDecoupledScalewise2024}.
Inspired by Speculative Decoding~\cite{leviathan2023fast}, CoDe~\cite{chenCollaborativeDecodingMakes2024} uses a large and a small model to reduce redundant computations on larger scales.
FastVAR~\cite{guoFastVARLinearVisual2025} and SparseVAR~\cite{chenFrequencyAwareAutoregressiveModeling2025} leverage the frequency characteristics of scale prediction to reduce computation by lowering the coefficient $\alpha$ in the time complexity $O(\alpha\times n^4)$, and SkipVAR~\cite{liSkipVARAcceleratingVisual2025} further explores this by exploiting high-frequency differences across different samples.
Architectural improvements to VAR, such as HMAR~\cite{kumbongHMAREfficientHierarchical} and HART~\cite{tangHARTEfficientVisual2024a}, require extra training or structural changes.
To the best of our knowledge, the closely related FreqExit~\cite{lifreqexit} is a training-time early-exiting method. In contrast, our approach is training-free and dynamically adjusts the network depth token-wise during inference.

\begin{figure}[t]
  \centering
  \begin{subfigure}[b]{0.455\linewidth}
    \includegraphics[width=\linewidth, trim=10 10 10 10, clip]{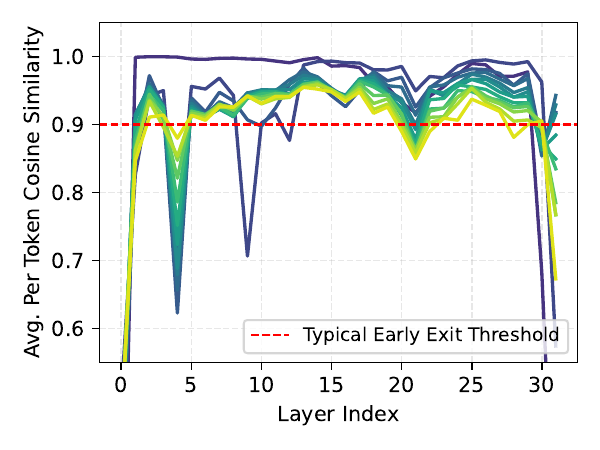}
    \caption{Token-wise Layer Similarity}
    \label{fig:layer_simularity}
  \end{subfigure}
  \begin{subfigure}[b]{0.515\linewidth}
    \includegraphics[width=\linewidth, trim=7 7 7 7, clip]{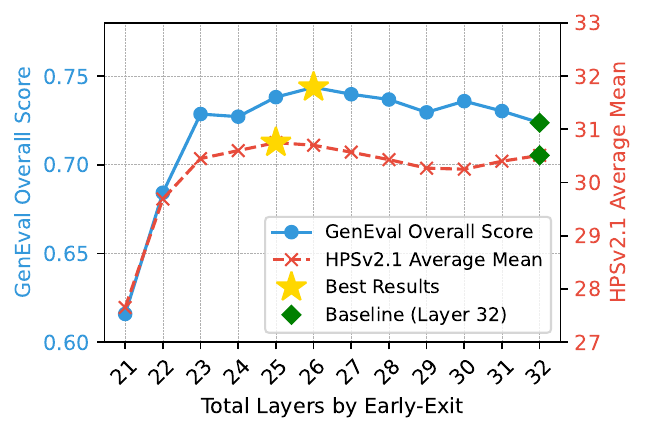}
    \caption{Early-exiting Evaluations}
    \label{fig:depth_redundancy}
  \end{subfigure}
    \caption{\textbf{Evidence of depth redundancy in pretrained VAR models.} (a) Token-wise representation similarity between consecutive layers shows saturation at different depths (darker colors for smaller scales). (b) Generation quality peaks before the final layer with early exiting, confirming full depth is not always optimal.}
  \label{fig:redundancy}
  \vspace{-1em}
\end{figure}

\section{Methodology}

\subsection{Empirical Observations}

\noindent\textbf{Limitations in Frequency Approximation.} 
Previous works~\cite{guoFastVARLinearVisual2025, chenFrequencyAwareAutoregressiveModeling2025} accelerate inference by pruning low-frequency tokens identified via approximated frequency maps, based on the observation that different frequency components converge at different rates during generation.
Contrary to intuition, our empirical analysis shows that more accurate frequency map approximations do not guarantee better image quality. On hard-pruning~\cite{chenFrequencyAwareAutoregressiveModeling2025}, we evaluated 800 samples by comparing its predicted frequency mask with a Sobel-filtered ground-truth. 
As shown in \cref{fig:freq_approx}, the approximation accuracy exhibits only a weak positive correlation with final image quality (SSIM; Pearson’s r = 0.138), indicating that refining frequency maps does not reliably improve results.
To further investigate, we conducted an oracle experiment using a perfect frequency mask derived directly from the ground-truth image. As shown in \cref{fig:freq_oracle}, even this ideal hard-pruning degrades quality, revealing a deeper issue with the assumption that low-frequency regions can be safely omitted.
These observations suggest that binary hard-pruning is inherently limited, motivating a shift towards more granular computational allocation strategies.

\noindent\textbf{Depth Redundancy in Pretrained VAR.} 
\label{sec:redundancy}
To improve generalization and prevent overfitting, VAR models~\cite{tianVisualAutoregressiveModeling,hanInfinityScalingBitwise2025,tangHARTEfficientVisual2024a} often employ LayerDrop~\cite{fanReducingTransformerDepth2019a} during training. This regularization strategy introduces depth redundancy~\cite{fanReducingTransformerDepth2019a, elhoushi2024layerskip} into the pretrained model, which can be leveraged for faster inference.
To confirm this redundancy, we follow~\cite{hanInfinityScalingBitwise2025} and measure performance when forcing all tokens to exit at earlier layers. As shown in \cref{fig:depth_redundancy}, generation quality on two benchmarks peaks before the final layer rather than increasing monotonically, indicating that the model is over-parameterized in depth and can be accelerated by trimming layers without harming quality.
We further examine token-wise similarity across consecutive layers (\cref{fig:layer_simularity}) and observe that token representations saturate at some layers, revealing token-specific depth redundancy.
This contrasts with prior findings~\cite{lifreqexit} based on cosine-similarity saturation used in classical early-exit methods~\cite{schusterConfidentAdaptiveLanguage2022}, suggesting that VAR models exhibit a distinct form of redundancy that may require tailored exploitation strategies.
These observations imply that simpler tokens may not benefit from full-depth processing, motivating our approach to adaptively reduce layer depth to better harness this property.

\begin{figure*}[t]
  \centering
  \includegraphics[width=\linewidth, trim=5 5 5 3, clip]{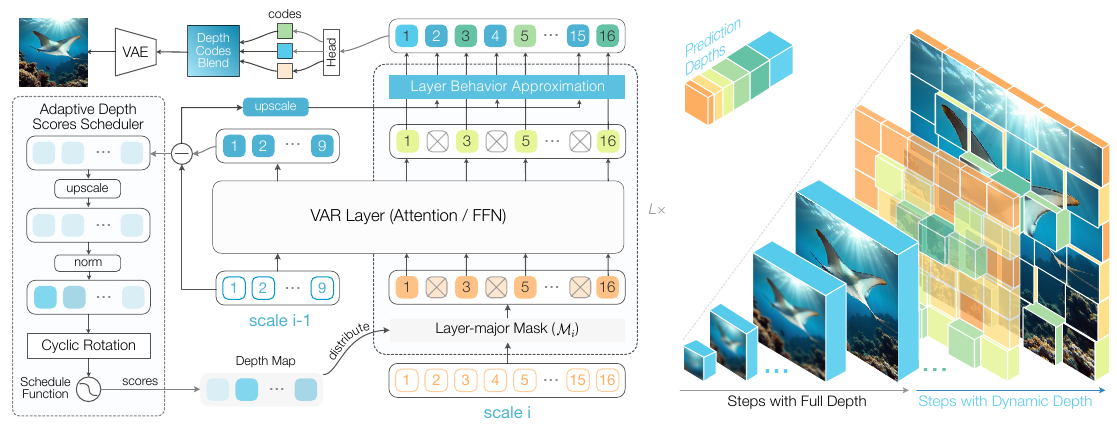}
  \caption{\textbf{Overview of dynamic depth inference in DepthVAR.} \textbf{Left:} At each scale $i$, we first use Adaptive Depth Score Scheduler to generate adaptive depth scores $\mathcal S_i$ using layer-wise changes from the previous scale, which are converted to a layer-major mask $\mathcal M_i$ via bit-reversal. The dynamic depth inference performs masked prediction using $\mathcal M_i$, reinstating cached layer behaviors from the last scale, and blends the resulting codes based on $\mathcal S_i$ to produce each scale’s final output. \textbf{Right:} Early steps follow standard inference, while dynamic depth is applied in later stages.}
  \label{fig:main}
  \vspace{-1em}
\end{figure*}

\subsection{Preliminaries}

At a high level, the visual autoregressive modeling (VAR)~\cite{tianVisualAutoregressiveModeling,hanInfinityScalingBitwise2025,tangHARTEfficientVisual2024a} predicts $K$ multi-scale residual maps $(r_0, r_1, \dots, r_{K-1})$ and accumulates the upscaled residual maps %
to gain a feature map $f_i$ at each scale $i \in \{0,\dots, {K-1}\}$. With encoded start token $r_0$ from text prompts~\cite{hanInfinityScalingBitwise2025,tangHARTEfficientVisual2024a, niSentenceT5ScalableSentence2021}, given a VAR model with $L$ stacked transformer layers~\cite{vaswaniAttentionAllYou2017}, the parallel prediction of $h_i\times w_i$ tokens in $r_i \in [V]^{h_i\times w_i}$ starts from scale-conditioned downsampled feature embeddings:
\begin{equation}
    r_i = \text{Layer}_{0\dots L}(\text{embed}(\textit{down}(f_{i-1})))
\end{equation}
where $\textit{down}$ is bilinear downsampling~\cite{hanInfinityScalingBitwise2025} and $\text{Layer}_{0\dots L}$ is the cumulative form of using $r_i^\ell=\text{Layer}_\ell(r_i^{\ell-1})$, where $\text{Layer}_\ell$ is the $\ell$-th block. 
Upsampling retrieved features $z_i$ from the codebook at the predicted logits $p_i=\text{head}(r_i)$ gives the intermediate feature maps:
\begin{equation}
    f_i = f_{i-1} + \textit{up}(z_i, h_{K-1}, w_{K-1})
    \end{equation}
where $z_i=\text{lookup}(p_i)$. In the last prediction step, $f_{K-1}$ is used for generating the prediction image. This standard prediction process utilizes all model layers and naturally allocates equal computation for each position on the image.

\subsection{DepthVAR: Predicting with Dynamic Depth}

As shown in \cref{fig:main}, our key intuition is to adaptively reduce computational depth by assigning each token map position a specific score, which are translated into layer-major masks to selectively activate transformer blocks, while masked positions are restored and resulting codes are blended to ensure each token's influence is proportional to its processing depth. Specifically at scale $i$, given precomputed depth scores $\mathcal S_i \in [0,1]^{h_i\times w_i}$ (\cref{sec:scheduler}), we first obtain the integer depth map $\mathcal D_i=\lfloor \mathcal S_i\cdot L       \rfloor$ to guide the mapping process.

\noindent\textbf{Depth Map to Layer-major Mask by Bit-reversal.} We then permute position-wise depths $\mathcal D_i \in \mathbb{N}^{h_i\times w_i}$ to layer-major masks $\mathcal{M}_i\in \{0,1\}^{L\times h_i\times w_i}$ for easier computation, by distributing per-token depths uniformly across layers for each position $(m,n)$ to achieve unbiased layer utilization. This prevents layers of shallower tokens from being disproportionately pruned. Let $k=\lceil\log_2 L\rceil$, and we define the bit-reversal permutation $\pi_L:\{0,\dots,L-1\}\to\{0,\dots,L-1\}$ by:
\begin{equation}
\pi_L(x) \;=\; \texttt{rev}_k(x)
\end{equation}
where
$x=\sum_{j=0}^{k-1} b_j 2^j, \texttt{rev}_k(x)=\sum_{j=0}^{k-1} b_j 2^{k-1-j}$. Bit-reversal is also the index ordering used in radix-2 Cooley–Tukey FFTs~\cite{cooley1965algorithm, elster1989fast}. For each token index $(m,n) \in [0, h_i)\times[0, w_i) \cap \mathbb{N}^2$, we choose the set of active layers as
$\mathcal L_i(m,n) \;=\; \{\,\pi_L(x)\,\}_{x=0}^{\,d_i(m,n)-1}$ and define the layer-major binary mask $\mathcal M_i\in\{0,1\}^{L\times h_i\times w_i}$ as:
\begin{equation}
\mathcal M_i(\ell,m,n)=\mathbf 1\!\left\{\ell\in \mathcal L_i(m,n)\right\}.
\end{equation}
For example, with $L=32$ and $d_i(m,n)=5$, one obtains $\mathcal L_i(m,n)=\{0,16,8,24,4\}$. 

\noindent\textbf{Masked Layer Behavior Approximation.} At each layer $\ell$, we reduce computation by processing only the active positions defined by the spatial mask slice $\mathcal M_i(\ell)$, and restore the cached proxy from the last scale at masked positions:
\begin{equation}
\label{eq:mask_pred}
\begin{split}
r_i^\ell = &\underbrace{\textbf{Layer}_\ell(r_i^{\ell-1}\odot \mathcal M_i(\ell))}_{\text{sparse prediction}}+ \\ 
& \underbrace{\textit up(r_{i-1}^{\ell}-r_{i-1}^{\ell-1}, h_i, w_i)\odot (1-\mathcal M_i(\ell))}_{\text{cached proxy restoration}}
\end{split}
\end{equation}
where $\textit up(\cdot, \cdot, \cdot)$ is the bilinear upscale. 
\cref{eq:mask_pred} ensures that subsequent layers receive a spatially complete feature map, allowing the upscaled residuals cached from the previous scale to serve as a consistent and continuous layer behavior proxy for masked regions.
We use the original positions $(m,n)$ as embedding positions in each layer block's RoPE2d~\cite{heoRotaryPositionEmbedding2025, hanInfinityScalingBitwise2025}, and restore $(1-\mathcal M_i(0))$ by the similarity criteria following~\cite{chenFrequencyAwareAutoregressiveModeling2025} after the last layer $\ell=L-1$, empirically minimizing the impact of masked tokens omitted from the attention context by exploiting inter-scale local stability.

\noindent\textbf{Depth-based Code Blending.} Finally, to let the residual added to the intermediate feature map be affected more by deeper tokens and less by shallower ones, we reweight the predicted codes 
$z_i$ by the depth scores map $\mathcal S_i =\big[\big[s_i(m,n)\big]_{m=0}^{h_i-1}\big]{}_{n=0}^{w_i-1}$ as $z_i=\mathcal S_i \cdot \text{lookup}(p_i)$, where $p_i=\text{head}(r_i^L)$. 
This ensures that the contribution of each token is proportional to its computational investment. %

\begin{table*}[ht]
  \caption{Quantitative evaluation on GenEval. Our DepthVAR achieves a superior trade-off between semantic consistency and inference latency compared to the baseline and other acceleration methods. ${}^\ddag$: requires additional training; $^{*}$: w/o last two scales; EE: early exit.}
    \label{tab:geneval_table}
    \centering
    \renewcommand{\arraystretch}{0.95} %
    \setlength{\tabcolsep}{5pt}
    \small
    \begin{tabular}{
      @{}c c c c c c c c c c @{}
    }
      \toprule

      \multirow{2}{*}{\textbf{Methods}} &
      \multicolumn{7}{c}{\textbf{GenEval} $\uparrow$} &
      \multirow{2}{*}{\textbf{\makecell{Avg\\Lat.(ms) $\downarrow$}}} &
      \multirow{2}{*}{\textbf{\makecell{Acc.\\Steps}}} 
      \\

      \cmidrule(lr){2-8}
      & \textbf{Counting} & \textbf{Color Attr.} &
        \textbf{Two Obj.} & \textbf{Colors} & \textbf{Position} & \textbf{Sin Obj.} & \textbf{Overall} & \\
      \midrule
      Infinity~\cite{hanInfinityScalingBitwise2025}        & 0.6812 & 0.5375 & 0.8636 & 0.8298 & 0.4300 & 1.0000 & 0.7237 & 2706& / \\
      Infinity\textit{-EE-26}~\cite{hanInfinityScalingBitwise2025} & 0.6875  & 0.5925 &  0.8561 & 0.8511 & 0.4750 & 1.0000 & 0.7437 & 2232 & 0-12 \\
      \hdashline
      + ToMe \{0.5, 0.5\}~\cite{bolya2022tome} & 0.6562 & 0.4050 & 0.7854 & 0.7287 & 0.4100 & 1.0000 & 0.6642 & 1284 & 11-12\\
      + SparseVAR-0.7~\cite{chenFrequencyAwareAutoregressiveModeling2025}& 0.6812 & 0.5500 & 0.8131 & 0.8378 & 0.4425 & 1.0000 & 0.7208 & 1281 & 10-12 \\
      + SkipVAR${}^\ddag$@0.84~\cite{liSkipVARAcceleratingVisual2025}& 0.7188 & 0.5375 & 0.8460 & 0.8431 & 0.3975 & 1.0000 & 0.7238 & 1325 & 10-12 \\
      + FastVAR$^{*}$~\cite{guoFastVARLinearVisual2025}& 0.7000 & 0.5525 & 0.8359 & 0.8245 & 0.4300 & 1.0000 & 0.7238 & 1080 & 9-10 \\
      \hdashline
      \rowcolor{gray!10} + DepthVAR$^{*}$($\mathcal{R}$=7)& 0.7188 & 0.5600 & 0.8157 & 0.8245 & 0.4050 & 1.0000 & 0.7207 & 869 & 8-10 \\
      \rowcolor{gray!10} + DepthVAR($\mathcal{R}$=7)& 0.6906 & 0.5575 & 0.8359 & 0.8271 & 0.4425 & 1.0000 & 0.7256 & 1168 & 8-12\\
      \rowcolor{gray!10} + DepthVAR($\mathcal{R}$=8)& 0.6812 & 0.5725 & 0.8333 & 0.8324 & 0.4375 & 1.0000 & 0.7262 & 1295 & 9-12 \\
      \rowcolor{gray!10} + DepthVAR($\mathcal{R}$=9)& 0.6969 & 0.5700 & 0.8333 & 0.8457 & 0.4450 & 1.0000 & 0.7318 & 1622 & 10-12 \\
      \midrule 
      HART\cite{tangHARTEfficientVisual2024a} & 0.3594 & 0.2550 & 0.7197 & 0.8644 & 0.1475 & 0.9875 & 0.5556 & 1102 & / \\
      HART\textit{-EE-21}\cite{tangHARTEfficientVisual2024a} & 0.3688 & 0.2550 & 0.7803 & 0.8590 & 0.1575 & 0.9750 & 0.5659 & 987 & 0-13 \\
      \hdashline
      + SparseVAR-0.7\cite{chenFrequencyAwareAutoregressiveModeling2025} & 0.3625 & 0.2200 & 0.6364 & 0.8484 & 0.1275 & 0.9750 & 0.5283 & 636 & 10-13 \\
      + FastVAR \textit{w/o FlashAttn}\cite{guoFastVARLinearVisual2025} & 0.3688 & 0.2175 & 0.6995 & 0.8378 & 0.1200 & 0.9750 & 0.5364 & 1195 & 12-13 \\
      \hdashline
      \rowcolor{gray!10} + DepthVAR($\mathcal{R}$=9) & 0.3656 & 0.2525 & 0.6869 & 0.8457 & 0.1450 & 0.9781 & 0.5456 & 710 & 10-13 \\
      \rowcolor{gray!10} + DepthVAR($\mathcal{R}$=10) & 0.3906 & 0.2750 & 0.7197 & 0.8697 & 0.1425 & 0.9875 & 0.5642 & 856 & 11-13 \\
      \bottomrule
    \end{tabular}
    \vspace{-1em}
\end{table*}

\begin{figure}[bt]
    \centering
    \includegraphics[width=\linewidth]{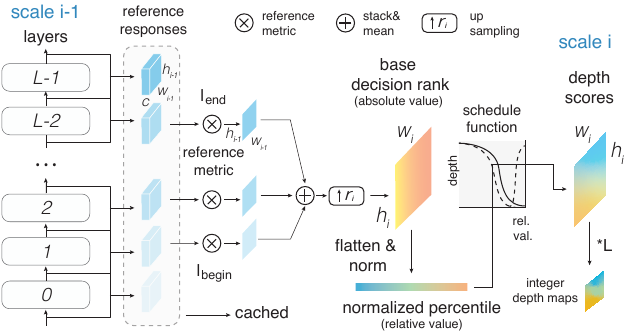}
    \caption{\textbf{Adaptive Depth Score Scheduler Pipeline.} Feature changes from scale $i-1$ are aggregated, upsampled, and normalized into percentiles, which are then mapped via a schedule function to continuous depth scores for scale $i$.
    }
    \label{fig:scheduler-pipe}
    \vspace{-1.1em}
\end{figure}

\subsection{Adaptive Depth Scores Scheduler}
\label{sec:scheduler}
This dynamic depth inference requires depth scores $\mathcal S_i\in[0,1]$ estimated for each token position. Directly porting language modeling~\cite{schusterConfidentAdaptiveLanguage2022} confidence scores is non-trivial: The visual code space is too large for reliable top-k softmax estimation; regions may visually saturate while hidden-state similarities still vary; and additional classifiers require extra fine-tuning. Prior approaches~\cite{guoFastVARLinearVisual2025,chenFrequencyAwareAutoregressiveModeling2025} approximate frequency rank maps either via MSE~\cite{chenFrequencyAwareAutoregressiveModeling2025} or mean-subtraction~\cite{guoFastVARLinearVisual2025}. However, as we showed in \cref{fig:freq_limit}, the issue lies more in how these ranks are used rather than the precision of the frequency approximation itself. %

Hence, we propose to interpret the previous $(i-1)$-th scale's layer-wise changes as `past decisions' that guides the current refinement. As illustrated in \cref{fig:scheduler-pipe}, this process involves aggregating absolute feature responses into a decision rank map, normalizing these into relative percentiles, and applying a schedule function to map importance to depth. Specifically, given a reference metric $\mathcal E$ and a reference range $[\ell_{\text{begin}}, \ell_{\text{end}}]$, we aggregate and upsample reference responses to form a \textbf{base decision rank map}: 
\begin{equation}
\mathcal B_i=up\big(\sum_{\ell=\ell_{\text{begin}}}^{\ell_{\text{end}}} \mathcal E(r_{i-1}^{\ell}-r_{i-1}^{\ell-1}), h_i, w_i\big) \in \mathbb{R}^{h_i\times w_i},
\end{equation}
and compute normalized decision rank percentiles by 
    $\rho_i(m,n)=\frac{1}{h_i w_i}
\sum_{(p,q)}\mathbf 1\!\left\{\mathcal B_i(p,q)>\mathcal B_i(m,n)\right\}\in[0,1]$.
With a monotonically decreasing \textbf{schedule function} $\mathcal G:[0,1]\rightarrow[0,1]$, we can convert decision rank percentiles to decayed depth scores $s_i(m,n)=\mathcal G(\rho_i(m,n))$. To further generalize, we modify $\mathcal G$ with a cyclic percentile rotation of magnitude $\eta\in (0,1)$, 
\begin{equation}
    \mathcal G'(\rho)=\begin{cases}
        \mathcal G(\frac{\rho}{\eta}) & 0 \leq \rho \leq \eta, \\
        \mathcal G(\frac{1-\rho}{1-\eta}) & \eta < \rho \leq 1.
    \end{cases}
\end{equation}
which prevents repeatedly updating the same tokens, as previously lowest ranks are rotated away. By applying the rotated mapping on the normalized rank percentiles, we obtain the adaptive \textbf{depth scores} as 
    $\mathcal S_i=\mathcal G'(\rho_i)$.

This percentile-based adaptive scheduling resembles dynamic depth transformers~\cite{gravesAdaptiveComputationTime2017, elbayad2020depth, raoDynamicViTEfficientVision}, and we extend it across multiscale predictions. By aligning the integral area of $\mathcal G'$ on a reference scale $r_\mathcal R$ with index $\mathcal R$, we constrain computation allocated for larger scales and achieve earlier computation reduction than others. In practice, we employ parameter controlled functions $\mathcal G_{\text{sigmoid}} = \frac{1}{1+e^{k(x-c)}}$, $\mathcal G_{\text{linear-a}}=1-c\cdot x$ and $\mathcal G_{\text{linear-b}}=c(x-1)$, where $k$ is a user-defined parameter and $c$ is solved from a given integral area.  For the reference metric, we use $\mathcal E_{\text{MAE}} =|\cdot|, \mathcal E_{\text{MSE}}=|\cdot|^2$, and $\mathcal E_{\text{sub}}=\cdot -\text{mean}(\cdot)$, we find these metrics behave similarly, while $\mathcal E_{\text{MAE}}$ offering the best overall balance.

\begin{figure*}[tb]
  \centering
  \includegraphics[width=\textwidth, trim=20 40 0 0, clip]{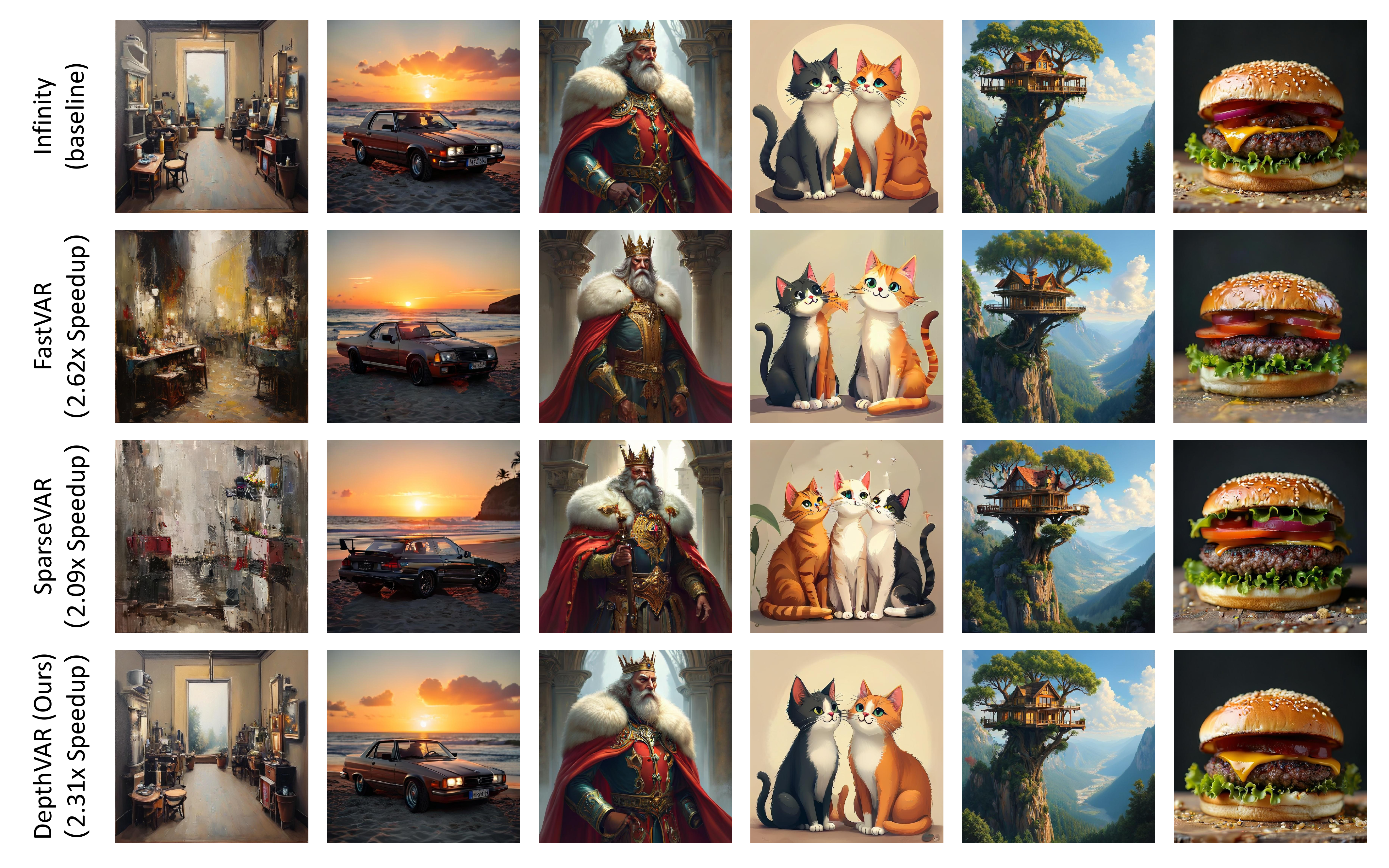}
  \caption{Qualitative visual comparisons between our method, the baseline, and other approaches with relatively fixed inference latency. Our method achieves a 2.31× acceleration while preserving semantic consistency and delivering rich visual details.}
  \label{fig:res}

\end{figure*}

\section{Experiments}
\subsection{Experimental Setup}
\noindent\textbf{Models and Metrics.} We apply DepthVAR on Infinity-2B~\cite{hanInfinityScalingBitwise2025} and HART-0.7B~\cite{tangHARTEfficientVisual2024a}, both are capable of generating images up to 1024×1024 high-resolution under text prompts. 
We evaluate them across multiple popular benchmarks: GenEval~\cite{ghosh2023geneval} for measuring high-level semantic consistency, while HPSv2.1~\cite{wu2023humanhps} and ImageReward~\cite{xu2023imagereward} for human-preference alignment. 
Unless otherwise specified, we conduct ablation studies on Infinity~\cite{hanInfinityScalingBitwise2025} for consistency and computational efficiency, while HART~\cite{tangHARTEfficientVisual2024a} is used solely for the main results. 
The performance of each acceleration method is quantified by reporting both its benchmark scores and its wall-clock runtime. To ensure a fair comparison, all methods' hyperparameters are kept as their default settings strictly for evaluation.

\begin{table*}[ht]
  \caption{Quantitative evaluation on Human Preference Metrics, including HPSv2.1 and ImageReward. DepthVAR demonstrates a strong balance between performance and efficiency.
  ${}^\ddag$: requires additional training; $^{*}$: w/o last two scales; EE: early exit.}
  \label{tab:hps_imagereward}
  \centering
  \renewcommand{\arraystretch}{0.95} %
  \setlength{\tabcolsep}{5pt}
  \small
  \begin{tabular}{@{}ccccccccc@{}}
    \toprule
    \multirow{2}{*}{\textbf{Methods}} & \multicolumn{6}{c}{\textbf{HPSv2.1}} & \multicolumn{2}{c}{\textbf{ImageReward}} \\
    \cmidrule(lr){2-7}\cmidrule(lr){8-9}
     & \textbf{Anime} & \textbf{Photo} & \textbf{Painting} & \textbf{Concept-Art} & \textbf{Overall$\uparrow$} & \textbf{\makecell{Latency (ms)$\downarrow$}} & \textbf{Score$\uparrow$} & \textbf{\makecell{Latency (ms)$\downarrow$}} \\
    \midrule
    Infinity~\cite{hanInfinityScalingBitwise2025}        & 31.68 & 29.39 & 30.44 & 30.36 & 30.47 & 2724 & 0.9515 & 2716 \\
    Infinity-\textit{EE-26}~\cite{hanInfinityScalingBitwise2025}        & 32.06 & 29.76 & 30.48 & 30.53 & 30.70 & 2210 & 0.8965 & 2206 \\
    \hdashline
    + ToMe \{0.5, 0.5\} & 28.65 & 26.46 & 27.12 & 27.10 & 27.33 & 1330 & 0.7840 & 1287 \\
    + SparseVAR-0.7~\cite{chenFrequencyAwareAutoregressiveModeling2025}& 31.03 & 28.74 & 29.57 & 29.68 & 29.76 & 1332 & 0.8936 & 1301 \\
    + SkipVAR${}^\ddag$@0.84~\cite{liSkipVARAcceleratingVisual2025}& 31.60 & 29.19 & 30.43 & 30.27 & 30.37 & 1692 & 0.9376 & 1744 \\
    + FastVAR$^{*}$~\cite{guoFastVARLinearVisual2025}& 31.08 & 28.82 & 29.97 & 29.86 & 29.93 & 1027 & 0.9116 & 1036 \\
    \hdashline
    \rowcolor{gray!10} + DepthVAR$^{*}$($\mathcal{R}$=7)& 31.20 & 28.95 & 29.94 & 29.85 & 29.98 & 882 & 0.8996 & 876 \\
    \rowcolor{gray!10} + DepthVAR($\mathcal{R}$=7)& 31.33 & 29.90 & 30.03 & 28.99 & 30.06 & 1185 & 0.9088 & 1174 \\
    \rowcolor{gray!10} + DepthVAR($\mathcal{R}$=8)& 31.42 & 29.97 & 30.15 & 29.10 & 30.16 & 1285 & 0.9171 & 1303 \\
    \rowcolor{gray!10} + DepthVAR($\mathcal{R}$=9)& 31.52 & 30.12 & 30.27 & 29.25 & 30.29 & 1625 & 0.9254 & 1616 \\
    \midrule
    HART~\cite{tangHARTEfficientVisual2024a} & 31.30 & 28.19 & 29.04 & 29.56 & 29.52 & 1109 & 0.9013 & 1103 \\
    HART\textit{-EE-21}~\cite{tangHARTEfficientVisual2024a} & 31.54 & 28.25 & 29.41 & 29.84 & 29.76 & 989 & 0.9004 & 985 \\
    \hdashline
    + SparseVAR-0.7~\cite{chenFrequencyAwareAutoregressiveModeling2025} & 27.69 & 25.33 & 25.60 & 26.12 & 26.18 & 669 & 0.5737 & 679\\
    + FastVAR \textit{w/o FlashAttn} ~\cite{guoFastVARLinearVisual2025} & 28.66 & 26.08 & 26.95 & 27.31 & 27.25 & 1208 & 0.7448 & 1209\\
    \hdashline
    \rowcolor{gray!10} + DepthVAR($\mathcal{R}$=9) & 28.44 & 25.72 & 26.44 & 26.63 & 26.81 & 729 & 0.6573 & 727\\
    \rowcolor{gray!10} + DepthVAR($\mathcal{R}$=10) & 29.95 & 27.04 & 27.72 & 28.10 & 28.20 & 885 & 0.7909 & 880 \\
    \bottomrule
  \end{tabular}
\vspace{-1em}
\end{table*}

\noindent\textbf{Implementation Details.} For the reference metric, we use MAE, with the reference range by $\ell_{\text{begin}}, \ell_{\text{end}}=3, 19$ for Infinity and $8, 17$ for HART. This range is chosen empirically with reference to layers that only attend to the foreground details~\cite{chenFrequencyAwareAutoregressiveModeling2025}.
We choose the sigmoid as the scheduling function $\mathcal G$ by setting $k=12$, and set $\eta=0.8$ for cyclic percentile rotation. We apply a similarity threshold of $0.9$ at a window of $5$ for the $1-\mathcal M_i(0)$ restoration.
For reference scale constrained compute, we align $\int_0^{\eta}\mathcal G'(\rho_i)d\rho_i, \forall i>\mathcal R$ to $\frac{h_\mathcal R \times w_\mathcal R}{h_i\times w_i}$, where $\mathcal R$ is the reference scale where later scales' computation is limited to, and we choose $i\in\{7,8,9\}$ for Infinity and $i\in\{9,10\}$ for HART. This enables DepthVAR to reduce computation earlier than other methods by up to two scales.
Following previous practices~\cite{guoFastVARLinearVisual2025, chenFrequencyAwareAutoregressiveModeling2025, liSkipVARAcceleratingVisual2025}, FlashAttn~\cite{dao2022flashattention} is applied to Infinity and not to HART, and the shared VAE latency cost is excluded from speed measurements. All experiments are conducted on a single NVIDIA RTX 3090 GPU with 24GB memory.

\subsection{Main Results}
\noindent\textbf{Comparison on GenEval.} 
We first evaluate DepthVAR across Infinity~\cite{hanInfinityScalingBitwise2025} and HART~\cite{tangHARTEfficientVisual2024a} on the GenEval benchmark~\cite{ghosh2023geneval}, comparing it with hard-pruning, token-merging, and early-exit baselines~\cite{guoFastVARLinearVisual2025, chenFrequencyAwareAutoregressiveModeling2025, bolya2022tome, hanInfinityScalingBitwise2025, tangHARTEfficientVisual2024a}.
As shown in \Cref{tab:geneval_table}, DepthVAR consistently achieves superior speed-quality trade-offs: on Infinity, it reaches a $2.3\times$-$3.1\times$ speedup (1168 or 869ms) with negligible quality loss, while on HART ($\mathcal R$=10), it improves the overall score by 1.5\% with $\sim1.3\times$ acceleration. Notably, while global early exit strategies~\cite{hanInfinityScalingBitwise2025, tangHARTEfficientVisual2024a} can improve scores, they offer limited 1.1$\times$-1.2$\times$ speedup. Unlike prior methods which suffer semantic instability at higher speedups, DepthVAR leverages exploitable depth redundancy (\cref{fig:depth_redundancy}) to maintain fidelity under constrained compute, and demonstrates competitive robustness and efficiency as a training-free framework.

\noindent\textbf{Comparison on HPSv2.1 and ImageReward.}
We evaluate human-preference alignment using HPSv2.1~\cite{wu2023humanhps} and ImageReward~\cite{xu2023imagereward}, with results presented in \Cref{tab:hps_imagereward}. DepthVAR shows a strong balance between quality and efficiency; on Infinity~\cite{hanInfinityScalingBitwise2025} it remains highly competitive with the baseline, while on HART~\cite{tangHARTEfficientVisual2024a}, it proves more robust without suffering dramatic score collapses. Overall, DepthVAR provides more flexible speed-quality trade-offs than SparseVAR~\cite{chenFrequencyAwareAutoregressiveModeling2025} and FastVAR~\cite{guoFastVARLinearVisual2025}, and offers a training-free solution with consistent runtime compared to SkipVAR~\cite{liSkipVARAcceleratingVisual2025} which requires extra training and has variable latency.

\noindent\textbf{Qualitative Visualizations.}
As shown in \Cref{fig:res}, we provide qualitative comparisons between DepthVAR, the baseline, and other acceleration methods at similar latency points. Our approach preserves high image quality and strong semantic consistency with the original generation, delivering rich visual details while achieving a 2.3$\times$ speedup. This visual evidence further demonstrates its superiority over existing hard-pruning approaches.

\begin{table}[tb]
  \caption{Ablation study of schedule functions on GenEval. Result scores are reported under compute constraints controlled by $R$.}
  \label{tab:schedule_score_latency}
  \centering
  \scalebox{0.9}{%
  \begin{tabular}{@{}c c c c@{}}
    \toprule
    Methods & \makecell{Schedule\\Function $\mathcal G$} & Score$\uparrow$ & \makecell{Avg\\Latency (ms)$\downarrow$} \\
    \midrule
    \multirow{5}{*}{\makecell{DepthVAR\\($\mathcal{R}$=7)}}
      & sigmoid@k=12 & 0.7256 & 1168 \\
      & sigmoid@k=3  & 0.7175 & 1139 \\
      & sigmoid@k=256 & 0.7212 & 1199 \\
      & linear-a      & 0.7218 & 1177 \\
      & linear-b      & 0.7250 & 1120 \\
    \midrule
    \multirow{5}{*}{\makecell{DepthVAR\\($\mathcal{R}$=8)}}
      & sigmoid@k=12 & 0.7262 & 1295 \\
      & sigmoid@k=3  & 0.7200 & 1285 \\
      & sigmoid@k=256& 0.7268 & 1314 \\
      & linear-a      & 0.7228 & 1316 \\
      & linear-b      & 0.7233 & 1284 \\
    \midrule
    \multirow{5}{*}{\makecell{DepthVAR\\($\mathcal{R}$=9)}}
      & sigmoid@k=12 & 0.7318 & 1622 \\
      & sigmoid@k=3  & 0.7316 & 1576 \\
      & sigmoid@k=256& 0.7310 & 1664 \\
      & linear-a      & 0.7310 & 1573 \\
      & linear-b      & 0.7291 & 1564 \\
    \bottomrule
  \end{tabular}
}
\vspace{-0.0em}
\end{table}

\begin{figure*}[ht]
  \centering
  \includegraphics[width=0.9\textwidth, trim=0 10 70 0, clip]{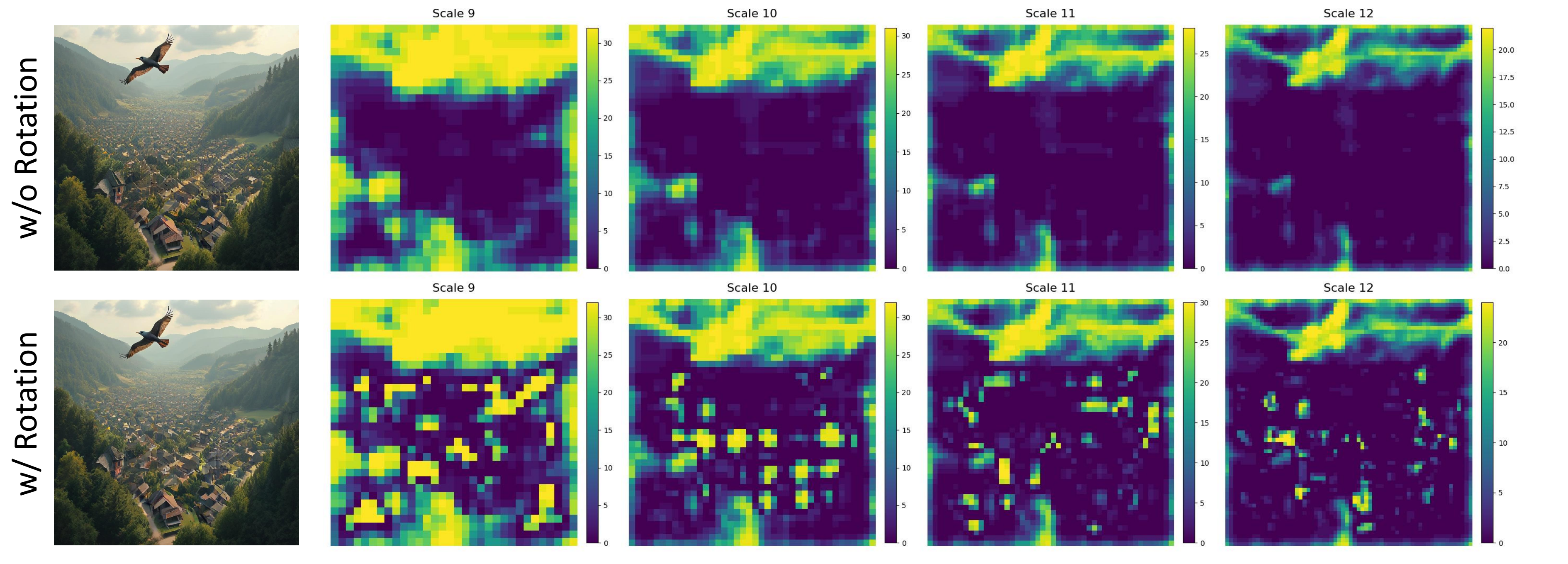}
  \caption{Visualization of depth maps in the presence and absence of Cyclic Percentile Rotation. This rotation operation enables updates in low-score regions that would otherwise remain unchanged. We present the visualizations of the last 4 scales (9-12) for clarity.}
  \label{fig:mask}
  \vspace{-0.0em}
\end{figure*}

\begin{figure*}[ht]
    \centering
    \begin{minipage}{0.25\textwidth}
            \centering
      \captionsetup{position=top} %
       \captionof{table}{Ablation study on the impact of Depth-based Code Blending. Disabling it degrades performance on GenEval.}
       \vspace{-0.3em}
        \resizebox{\linewidth}{!}{%
        \setlength{\tabcolsep}{4pt}
        \begin{tabular}{@{}lcc@{}}
          \toprule
          \multirow{2}{*}{\quad Methods} & \multicolumn{2}{c}{GenEval} \\
          \cmidrule(lr){2-3}
          & Score$\uparrow$ & Latency (ms)$\downarrow$ \\
          \midrule
          DepthVAR($\mathcal{R}$=7)                 & 0.7256 & 1168 \\
          \quad\textit{w/o Blend.}    & 0.7102 & 1172 \\
          DepthVAR($\mathcal{R}$=8)                 & 0.7262 & 1295 \\
          \quad\textit{w/o Blend.}    & 0.7242 & 1294 \\
          DepthVAR($\mathcal{R}$=9)                 & 0.7318 & 1622 \\
          \quad\textit{w/o Blend.}    & 0.7241 & 1625 \\
          \bottomrule
        \end{tabular}
      }%
      \label{tab:geneval_codes_blend}
    \end{minipage}
    \hfill
    \begin{minipage}{0.4\textwidth}
            \centering
      \captionof{table}{Ablation study on Cyclic Percentile Rotation. The table reports results with and without this component, illustrating its impact on overall performance.}
      \resizebox{\linewidth}{!}{%
        \setlength{\tabcolsep}{4pt}
        \begin{tabular}{@{}l cc cc@{}}
          \toprule
          \multirow{2}{*}{\quad Methods} & \multicolumn{2}{c}{GenEval} & \multicolumn{2}{c}{ImageReward} \\
          \cmidrule(lr){2-3}\cmidrule(lr){4-5}
          & Score$\uparrow$ & \makecell{Latency (ms)$\downarrow$} & Score$\uparrow$ & \makecell{Latency (ms)$\downarrow$} \\
          \midrule
          DepthVAR($\mathcal{R}$=7)        & 0.7256 & 1168 & 0.9088 & 1174 \\
          \quad\textit{w/o Rotation}  & 0.7277 & 1088 & 0.9031 & 1096 \\
          \addlinespace[2pt]
          DepthVAR($\mathcal{R}$=8)        & 0.7262 & 1295 & 0.9171 & 1303 \\
          \quad\textit{w/o Rotation}  & 0.7222 & 1206 & 0.9133 & 1231 \\
          \addlinespace[2pt]
          DepthVAR($\mathcal{R}$=9)        & 0.7318 & 1622 & 0.9254 & 1616 \\
          \quad\textit{w/o Rotation}  & 0.7274 & 1545 & 0.9231 & 1546 \\
          \bottomrule
        \end{tabular}
      }%
      \label{tab:ablate-cpr}
    \end{minipage}
    \hfill
    \begin{minipage}{0.3\textwidth}
            \centering
      \includegraphics[width=0.9\linewidth, , trim=10 10 0 0, clip]{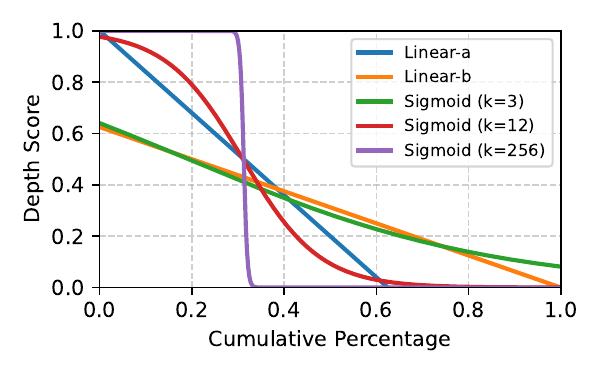}
      \caption{Profiles of different schedule functions under the same constraint. Linear $a$ and $b$ pass $(0, 1)$ and $(1,0)$.}
      \label{fig:curve}
    \end{minipage}
    \vspace{-1em}
\end{figure*}

\subsection{Ablation Studies}

\noindent\textbf{Impact of Depth-based Code Blending.}
To investigate whether tokens from deeper or shallower layers should contribute differently to image refinement, we conduct ablation experiments on the codes blending strategy across different reference scales.
As shown in \cref{tab:geneval_codes_blend}, with depth-based codes blending disabled\textemdash i.e., when the current-scale codes residual prediction is directly added to the feature map\textemdash the model performance drops by 0.002-0.015. 
Rather than contributing equally, this result suggests that deeper tokens should drive the primary updates, while shallower tokens provide fine-grained adjustments.

\noindent\textbf{Choice of Schedule Functions.}
The choice of the schedule function $\mathcal G$ determines how computational depth is distributed across tokens. To investigate the impact of different allocation strategies, we experiment with several functional forms for $\mathcal G$. As illustrated in Figure \ref{fig:curve}, these functions create distinct profiles for assigning layer depths based on token rank, ranging from gradual tapering (sigmoid) to linear decay.
We evaluate their effectiveness by comparing GenEval scores across different compute constraints (Table \ref{tab:schedule_score_latency}). The sigmoid function ($\mathcal G_{\text{sigmoid}}$ with $k=12$) almost consistently achieves the best performance, validating our approach of assigning varied layer depths across tokens. 
It also suggests that it is not necessary to enforce that all tokens are processed (as in $\mathcal G_{\text{linear-b}}$) or that some tokens always traverse the full network depth ($\mathcal G_{\text{linear-a}}$). 
This validates our continuous compute allocation strategy, demonstrating the effectiveness of a balanced schedule profile.

\noindent\textbf{Impact of Cyclic Percentile Rotation.}
Cyclic percentile rotation plays a critical role in re-ranking tokens for mapping depth scores, breaking the self-reinforcing top selection pattern that often emerges in hard-pruning.
An ablation study in \cref{tab:ablate-cpr} shows that incorporating cyclic percentile rotation consistently yields quality gains.
Moreover, as illustrated in \cref{fig:mask}, applying it enables regions with originally low scores in the decision map to receive substantive updates. 
With compute constraints relative to reference scales, cyclic percentile rotation ensures that tokens across spatial regions are iteratively and more evenly updated, rather than focusing updates only on a small subset.

\raggedbottom
\section{Conclusion}
In this paper, we introduce DepthVAR, a training-free framework that enables dynamic and continuous computational allocation for each token in Visual Autoregressive models. We analyze the limitations of frequency-based assumptions in hard-pruning acceleration methods and identify exploitable depth redundancy in VAR models. To leverage this redundancy and overcome the pitfalls of binary pruning, DepthVAR employs an adaptive depth scheduler with a cyclic rotated schedule function to heuristically assign computational depth per token. This is realized through a dynamic inference process using a bit-reversal layer-major mask and depth-based code blending. Experiments validate that DepthVAR achieves a superior trade-off between inference latency and performance on various semantic and human preference benchmarks compared to prior hard-pruning approaches, confirming its effectiveness.

\noindent \textbf{Limitations and Future Work.} 
A limitation of our work is the fixed per-sample compute budget. Future work could explore dynamic total compute allocation via routing or early-exiting, and investigate more advanced strategies for exploiting depth redundancy.

\section*{Acknowledgments}
This work was supported in part by the Open Fund of the Zhongguancun Open Laboratory of Optoelectronic Measurement and Intelligent Perception under the project `Lightweight Algorithm Design for Multimodal Object Recognition on Spaceborne Platforms'.

{
    \small
    \bibliographystyle{ieeenat_fullname}
    \bibliography{main}
}

\clearpage
\setcounter{page}{1}
\maketitlesupplementary
\appendix
\section{Ablation on Different Reference Metrics}
The reference metric $\mathcal E$ and its layer range $[\ell_{\text{begin}}, \ell_{\text{end}}]$ determine the base decision rank map that guides depth allocation. We ablate these choices in \Cref{tab:refmetric_geneval_imreward}, including metrics analogous to those in SparseVAR~\cite{chenFrequencyAwareAutoregressiveModeling2025} ($\mathcal E_{\text{MSE}}$ on Block 3) and FastVAR~\cite{guoFastVARLinearVisual2025} ($\mathcal E_{\text{SUB}}$). 
While ($\mathcal E_{\text{MAE}}$, $[3,19]$) and ($\mathcal E_{\text{MSE}}$, $[0,31]$) are mostly comparable, the MAE metric offers a more balanced trade-off between quality and latency. 
This highlights that the effectiveness of our framework stems from how ranks are utilized for dynamic depth scheduling, rather than the precision of the initial rank estimation itself.

\begin{table}[ht]
  \centering
  \caption{Ablation study on reference metrics. We compare different metrics, including those analogous to SparseVAR ($\mathcal E_{\text{MSE}}$, Block 3) and FastVAR ($\mathcal E_{\text{SUB}}$), under different reference scales $r_\mathcal R$.}
  \label{tab:refmetric_geneval_imreward}
  \scalebox{0.8}{%
  \setlength{\tabcolsep}{4pt}
  \begin{tabular}{@{}c c c c c c c@{}}
    \toprule
    \multirow{2}{*}{$\mathcal R$} &
    \multicolumn{2}{c}{Reference Metric} &
    \multicolumn{2}{c}{GenEval} &
    \multicolumn{2}{c}{ImageReward} \\
    \cmidrule(lr){2-3}\cmidrule(lr){4-5}\cmidrule(lr){6-7}
    & $\mathcal{E}$ & $[\ell_{\text{begin}}, \ell_{\text{end}}]$
    & Score$\uparrow$ & \makecell{Avg\\Latency (ms)$\downarrow$}
    & Score$\uparrow$ & \makecell{Avg\\Latency (ms)$\downarrow$} \\
    \midrule

    \multirow{6}{*}{\makecell{$7$}}
      & $\mathcal E_{\text{MAE}}$ & $[3,19]$  & 0.7256 & 1168 & 0.9088 & 1174 \\
      & $\mathcal E_{\text{MAE}}$ & $[0,31]$  & 0.7216 & 1228 & 0.9081 & 1253 \\
      & $\mathcal E_{\text{MSE}}$ & $[3,19]$  & 0.7219 & 1217 & 0.9094 & 1214 \\
      & $\mathcal E_{\text{MSE}}$ & $[0,31]$  & 0.7304 & 1270 & 0.8948 & 1295 \\
      & $\mathcal E_{\text{MSE}}$ & Block 3  & 0.7198 & 1164 & 0.9078 & 1184 \\
      & $\mathcal E_{\text{SUB}}$ & $-$   & 0.7210 & 1242 & 0.9033 & 1230 \\
    \midrule

    \multirow{6}{*}{\makecell{$8$}}
      & $\mathcal E_{\text{MAE}}$ & $[3,19]$  & 0.7262 & 1295 & 0.9171 & 1303 \\
      & $\mathcal E_{\text{MAE}}$ & $[0,31]$  & 0.7215 & 1339 & 0.9148 & 1353 \\
      & $\mathcal E_{\text{MSE}}$ & $[3,19]$  & 0.7241 & 1336 & 0.9172 & 1348 \\
      & $\mathcal E_{\text{MSE}}$ & $[0,31]$  & 0.7300 & 1377 & 0.9108 & 1384 \\
      & $\mathcal E_{\text{MSE}}$ & Block 3  & 0.7226 & 1302 & 0.9155 & 1307 \\
      & $\mathcal E_{\text{SUB}}$ & $-$   & 0.7207 & 1371 & 0.9152 & 1392 \\
    \midrule

    \multirow{6}{*}{\makecell{$9$}}
      & $\mathcal E_{\text{MAE}}$ & $[3,19]$  & 0.7318 & 1622 & 0.9254 & 1616 \\
      & $\mathcal E_{\text{MAE}}$ & $[0,31]$  & 0.7198 & 1651 & 0.9292 & 1670 \\
      & $\mathcal E_{\text{MSE}}$ & $[3,19]$  & 0.7282 & 1660 & 0.9236 & 1663 \\
      & $\mathcal E_{\text{MSE}}$ & $[0,31]$  & 0.7310 & 1686 & 0.9231 & 1692 \\
      & $\mathcal E_{\text{MSE}}$ & Block 3  & 0.7276 & 1623 & 0.9271 & 1638 \\
      & $\mathcal E_{\text{SUB}}$ & $-$   & 0.7323 & 1663 & 0.9261 & 1689 \\
    \bottomrule
  \end{tabular}
}
\vspace{-1em}
\end{table}

\section{Sensitivity Analysis of Reference Ranges}

\begin{figure}[b]
  \centering
  \begin{minipage}{\linewidth}
    \centering
    \includegraphics[width=\linewidth]{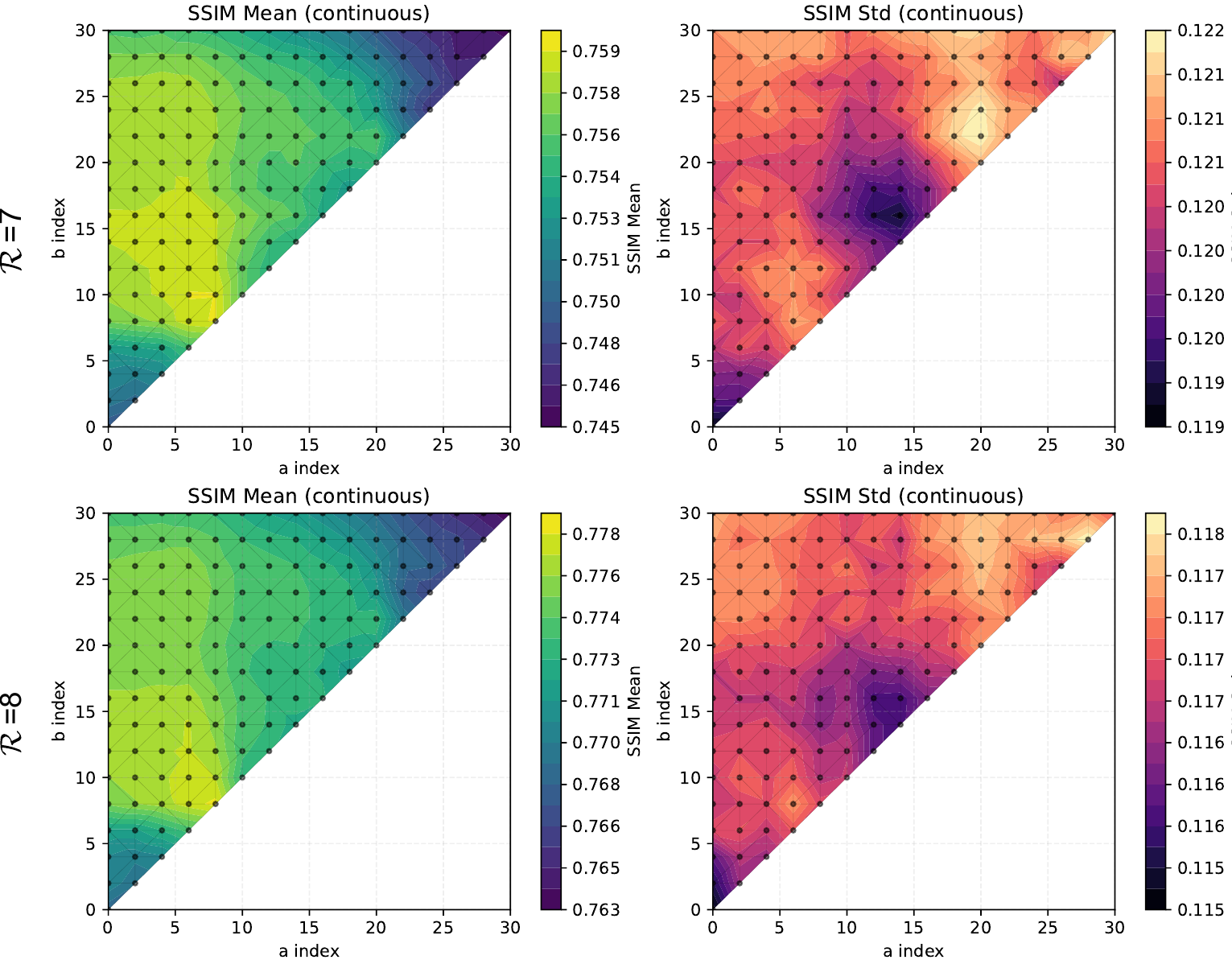}
    \caption{Sensitivity analysis of $[\ell_{\text{begin}}, \ell_{\text{end}}]=[a,b], \mathcal E=\mathcal E_{\text{MAE}}$. 
    }
    \label{fig:sensitivity_range_mae}
  \end{minipage}
  
  \vspace{0.5em} %
  
  \begin{minipage}{\linewidth}
    \centering
    \includegraphics[width=\linewidth]{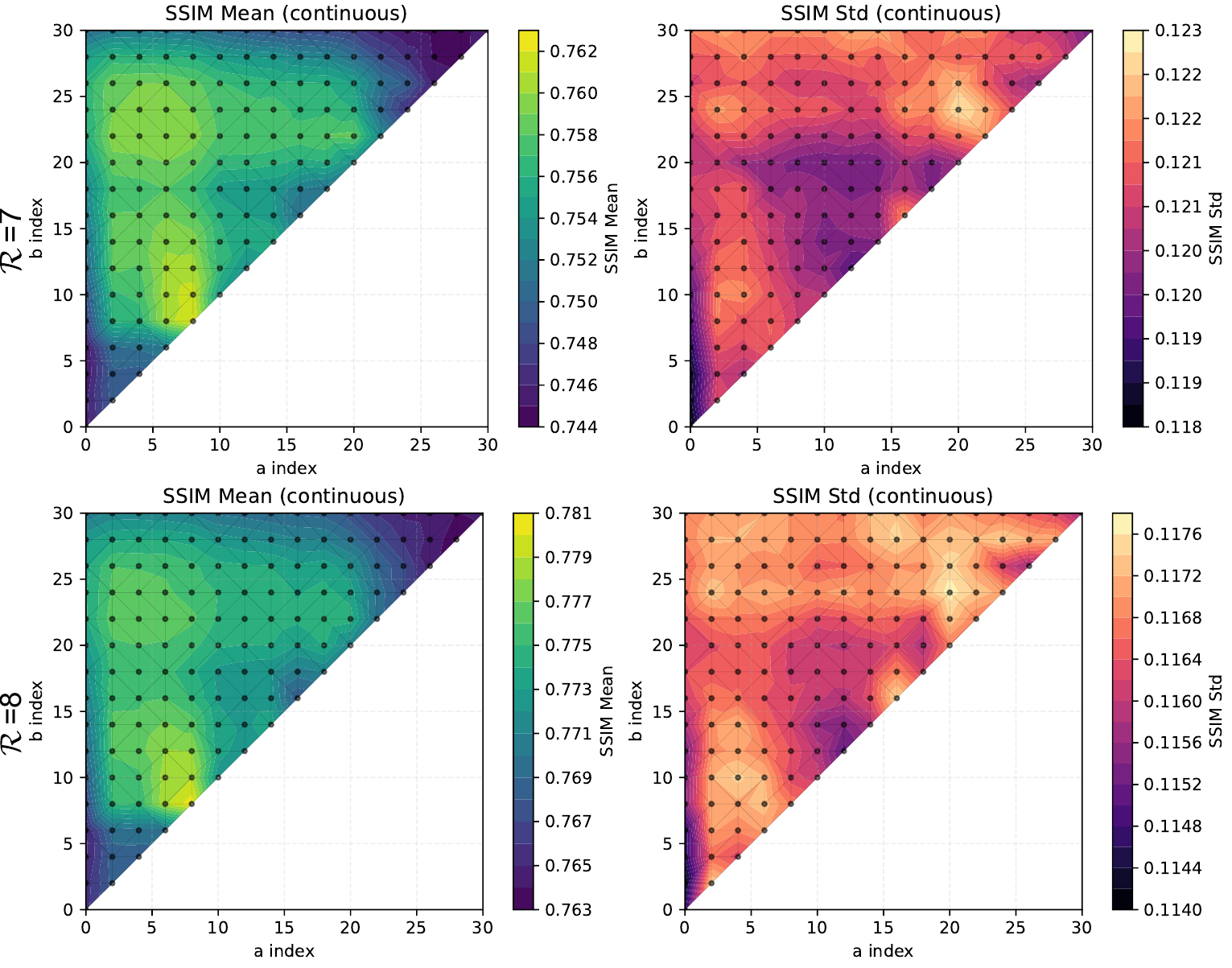}
    \caption{Sensitivity analysis of $[\ell_{\text{begin}}, \ell_{\text{end}}]=[a,b], \mathcal E=\mathcal E_{\text{MSE}}$. }
    \label{fig:sensitivity_range_mse}
  \end{minipage}
  \vspace{-0.8em}
\end{figure}

\begin{figure}[htb]
  \centering
  \begin{subfigure}[t]{0.51\linewidth}
    \includegraphics[width=\linewidth]{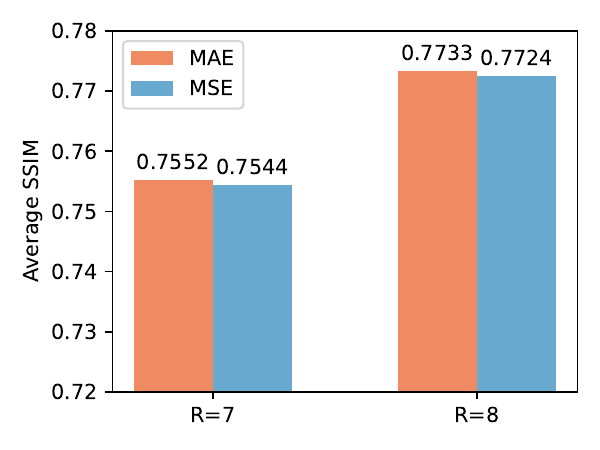}
    \caption{Comparison of mean SSIM averaged over ranges.}
    \label{fig:reference_metrics}
  \end{subfigure}
  \hfill
  \begin{subfigure}[t]{0.44\linewidth}
    \includegraphics[width=\linewidth]{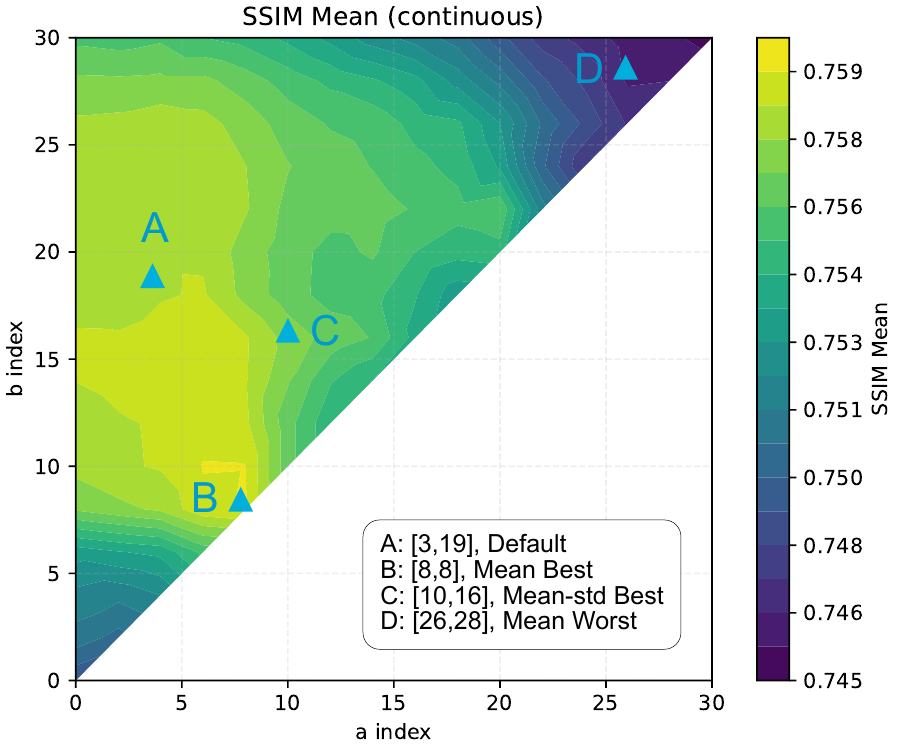}
    \caption{Selection of four layer range configurations.}
    \label{fig:layer_configuration}
  \end{subfigure}
  \caption{(a) Comparison of reference metrics, where $\mathcal{E}_{\text{MAE}}$ shows a marginally better average SSIM. (b) Selection of four specific layer configurations: (A) our default, (B) best SSIM mean, (C) best SSIM mean+std, and (D) near-worst SSIM.}
  \label{fig:sensitivity_combined}
\end{figure}

We conduct a sensitivity analysis to assess how the choice of the reference layer range $[\ell_{\text{begin}}, \ell_{\text{end}}]$ impacts generation quality. 
To map the optimization landscape, we compute the mean and standard deviation of SSIM scores on 100 prompts generated by expanding class names~\cite{openai2025gptoss120bgptoss20bmodel, Krizhevsky09learningmultiple},
for both MAE and MSE metrics, with a range step of 2. As illustrated in \cref{fig:sensitivity_range_mae,fig:sensitivity_range_mse}, the response patterns are consistent across different computational constraints ($\mathcal R=7,8$), indicating that the optimal layer range is independent of the overall compute budget. While MSE yields a marginally higher peak SSIM, MAE demonstrates a broader performance plateau, resulting in a slightly better average SSIM across all configurations (\cref{fig:reference_metrics}).

\begin{table}[tb]
  \centering
  \caption{Performance of selected layer range configurations on GenEval and ImageReward benchmarks. The configurations A, B, C, and D correspond to the points highlighted in \cref{fig:layer_configuration}.}
  \label{tab:layer_config_performance}
  \scalebox{0.9}{
    \setlength{\tabcolsep}{4pt}
  \begin{tabular}{@{}ccccc@{}}
    \toprule
    $\mathcal R$ & Selection & $[\ell_{\text{begin}}, \ell_{\text{end}}]$ & \makecell{GenEval \\ Overall}$\uparrow$ & \makecell{Image \\Reward}$\uparrow$ \\
    \midrule
    \multirow{4}{*}{7} & A & [3, 19] & 0.7256 & 0.9088 \\
                       & B & [8, 8]  & 0.7179 & 0.9096 \\
                       & C & [10, 16]& 0.7283 & 0.9101 \\
                       & D & [26, 28]& 0.7220 & 0.9016 \\
    \midrule
    \multirow{4}{*}{8} & A & [3, 19] & 0.7262 & 0.9171 \\
                       & B & [8, 8]  & 0.7260 & 0.9179 \\
                       & C & [10, 16]& 0.7313 & 0.9188 \\
                       & D & [26, 28]& 0.7259 & 0.9095 \\
    \bottomrule
  \end{tabular}
  }
\end{table}

To validate these observations on benchmarks, we select four representative configurations on MAE as highlighted in \cref{fig:layer_configuration}: (A) our default setting $[3, 19]$, (B) a single-layer reference $[8, 8]$ yielding the highest mean SSIM, (C) a statistical optimum $[10, 16]$ balancing mean and std, and (D) a sub-optimal region $[26, 28]$. \Cref{tab:layer_config_performance} reports their performance on GenEval~\cite{ghosh2023geneval} and ImageReward~\cite{xu2023imagereward}.
Notably, while the statistically optimal trade-off (C) yields the highest scores, sufficient baseline performance is maintained as long as the configuration resides within the stable plateau (setting A) and avoids poor choices like low-performing regions (setting D) or an insufficient number of reference layers (setting B). This observation, together with the SSIM sensitivity visualization (\cref{fig:sensitivity_range_mae}), highlights DepthVAR's low sensitivity to its hyperparameters, ensuring that suboptimal choices do not lead to catastrophic performance degradation.

\begin{table}[tb]
  \centering
  \caption{Performance comparison of bit-reversal and uniform configurations on GenEval, ImageReward, and HPSv2.1 benchmarks.}
  \label{tab:ablation_bitreversal}
  \scalebox{0.75}{
    \setlength{\tabcolsep}{2pt}
  \begin{tabular}{@{}cccccccccc@{}}
    \toprule
    \multirow{2}{*}{$\mathcal R$} & \multirow{2}{*}{\makecell{Configuration}} & \multicolumn{2}{c}{\makecell{GenEval}} & \multicolumn{2}{c}{\makecell{ImageReward}} & \multicolumn{2}{c}{\makecell{HPSv2.1}}\\
    \cmidrule(lr){3-4}\cmidrule(lr){5-6}\cmidrule(lr){7-8}
    & & \makecell{Overall $\uparrow$} & \makecell{Latency\\(ms)}$\downarrow$ & \makecell{Overall $\uparrow$} & \makecell{Latency\\(ms)}$\downarrow$ & \makecell{Score $\uparrow$} & \makecell{Latency\\(ms)}$\downarrow$ \\
    \midrule
    \multirow{2}{*}{7} & bit-reversal & 0.7256 & 1168 & 0.9088 & 1174 & 30.06 & 1185 \\
      & uniform      & 0.7255 & 1207 & 0.9070 & 1218 & 30.01 & 1232 \\
    \midrule
    \multirow{2}{*}{8} & bit-reversal & 0.7262 & 1295 & 0.9171 & 1303 & 30.16 & 1285 \\
      & uniform      & 0.7258 & 1346 & 0.9156 & 1356 & 30.13 & 1340 \\
    \midrule
    \multirow{2}{*}{9} & bit-reversal & 0.7318 & 1622 & 0.9254 & 1616 & 30.29 & 1625 \\
      & uniform      & 0.7263 & 1677 & 0.9291 & 1670 & 30.27 & 1667 \\
    \bottomrule
  \end{tabular}
  }
  \vspace{-0.8em}
\end{table}

\section{Ablation on Bit-reversal}
We validate the design of our layer selection mechanism by comparing our bit-reversal permutation against a uniform sampling strategy.
The uniform baseline distributes active layers equidistantly, whereas our bit-reversal method functions as a quasi-Monte Carlo sampler.
As shown in \Cref{tab:ablation_bitreversal}, the bit-reversal configuration consistently outperforms the uniform selection across all benchmarks and reference scale with slightly lower latencies, albeit with small margins. These results confirm the benefits of distributing active layers with bit-reversal.

\begin{figure}[tb]
  \centering
  \includegraphics[width=\linewidth]{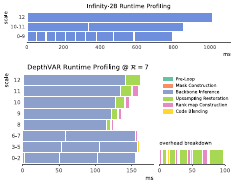}
  \caption{Runtime breakdown analysis on Infinity~\cite{hanInfinityScalingBitwise2025}.}
  \label{fig:runtime}
\end{figure}
\section{More Efficiency Profiling}

As illustrated in \cref{fig:runtime}, we profile the runtime overhead of DepthVAR on Infinity with $\mathcal{R}$=7. The breakdown shows that our acceleration framework introduces $\sim$100ms of additional overhead, dominated by rank-map construction and upsampling operations, which represent roughly 6\% of the total computation savings. Furthermore, caching intermediate layer behaviors from the previous scale incurs approximately 1.1GB of peak GPU memory overhead. The overall memory footprint is reduced \textbf{from 16.5GB to 11.6GB}, achieving 4.9GB savings and outperforming FastVAR~\cite{guoFastVARLinearVisual2025}'s 4.2GB reduction by an additional 700MB.

\begin{figure}[htb]
  \centering
  \begin{subfigure}[t]{0.48\linewidth}
    \includegraphics[width=\linewidth, trim=10 10 10 10, clip]{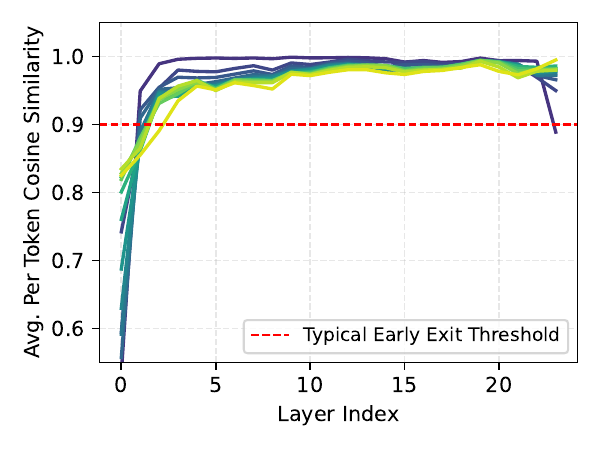}
    \caption{Token-wise Layer Similarity.}
    \label{fig:hart_layer_simularity}
  \end{subfigure}
  \hfill
  \begin{subfigure}[t]{0.45\linewidth}
    \includegraphics[width=\linewidth, trim=7 7 7 7, clip]{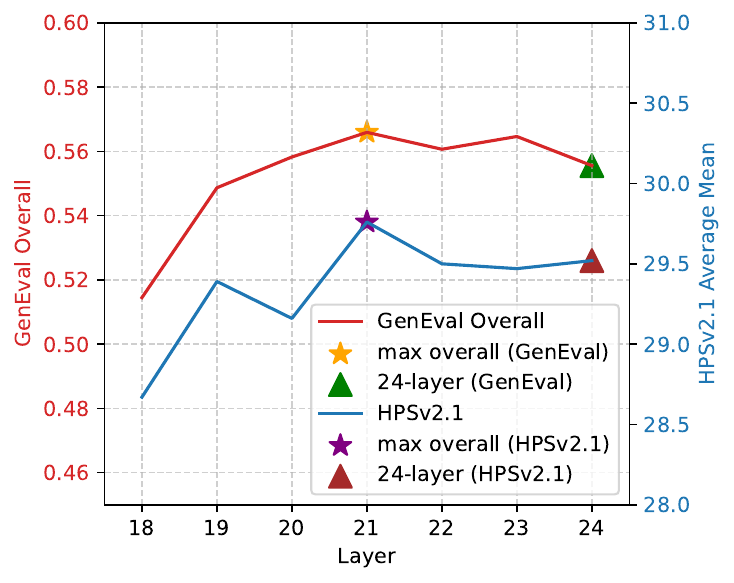}
    \caption{Early-exiting Evaluations.}
    \label{fig:hart_depth_redundancy}
  \end{subfigure}
    \caption{Generalization of depth redundancy to the HART~\cite{tangHARTEfficientVisual2024a} architecture. (a) Token-wise representation similarity exhibits saturation patterns consistent with standard VAR models. (b) Early-exiting evaluations on GenEval~\cite{ghosh2023geneval} and HPSv2.1~\cite{wu2023humanhps} confirm that generation quality peaks prior to the final layer.}
  \label{fig:hart_redundancy}
  \vspace{-0.8em}
\end{figure}

\section{Universality of Depth Redundancy}
To show that the observed depth redundancy is a fundamental property of VAR models, we extend our analysis from \cref{sec:redundancy} to the HART~\cite{tangHARTEfficientVisual2024a} architecture, a distinct hybrid variant. Applying the same evaluation protocols, we conducted token-wise layer similarity and early-exiting analyses. As shown in \cref{fig:hart_redundancy}, the results confirm that HART exhibits similar redundancy patterns, with generation quality peaking at 21 before the full layer. This suggests that depth redundancy is a pervasive characteristic of visual autoregressive models, validating the broader applicability of our adaptive depth paradigm to other architectures.

\begin{figure}[t]
  \centering
  \begin{minipage}[t]{0.55\linewidth}
    \centering
    \includegraphics[width=\linewidth]{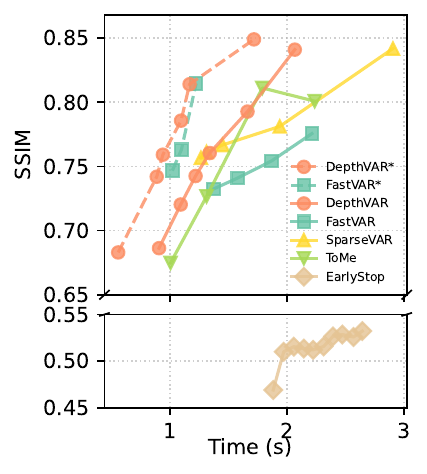}
    \caption{SSIM--latency pareto frontier. Dashed ${}^{*}$: w/o last 2 scales. DepthVAR consistently traces the upper-left.}
    \label{fig:time_vs_metrics}
  \end{minipage}\hfill
  \begin{minipage}[t]{0.43\linewidth}
    \centering
    \includegraphics[width=\linewidth]{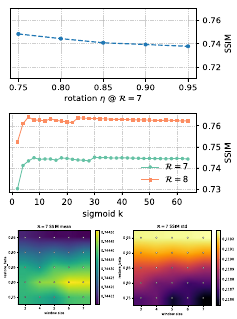}
    \caption{Sensitivity analysis over additional hyperparameters.}
    \label{fig:sensitivity}
  \end{minipage}
  \vspace{-1em}
\end{figure}

\section{Extended Pareto Frontier Analysis}
While individual points provide specific quality-speed results, our method's superiority is best characterized by the Pareto frontier, as in Fig.\ref{fig:time_vs_metrics}. We analyze the overall structural fidelity of DepthVAR on the Infinity~\cite{hanInfinityScalingBitwise2025} following the evaluation protocol in Appx.B. An observation of simple early-stopping truncation is that it significantly alters the generative trajectory, making it not a direct substitute if aiming to preserve structural identity. To ensure a fair comparison with FastVAR~\cite{guoFastVARLinearVisual2025}, we report both full-scale and omitting the final two scales. In both cases, DepthVAR traces the upper-left envelope, demonstrating the trade-off.

\section{More Hyperparameter Sensitivity}

We provide additional analysis for the remaining hyperparameters in Fig.~\ref{fig:sensitivity}, showing their properties. We observe that the rotation magnitude $\eta$ primarily balances overall structural quality against semantic alignment. For the schedule sharpness $k$, SSIM exhibits local peaks near 8, 18, and 30, making $k\in[8,30]$ a robust range for stable inference. Finally, minimal sensitivity to the restoration window size and $\beta$ is observed, suggesting potential simplification.

\section{More Qualitative Visualizations}
\begin{figure}[tb]
  \centering
  \begin{subfigure}[t]{0.85\linewidth}
    \includegraphics[width=\linewidth, trim=38 0 10 0, clip]{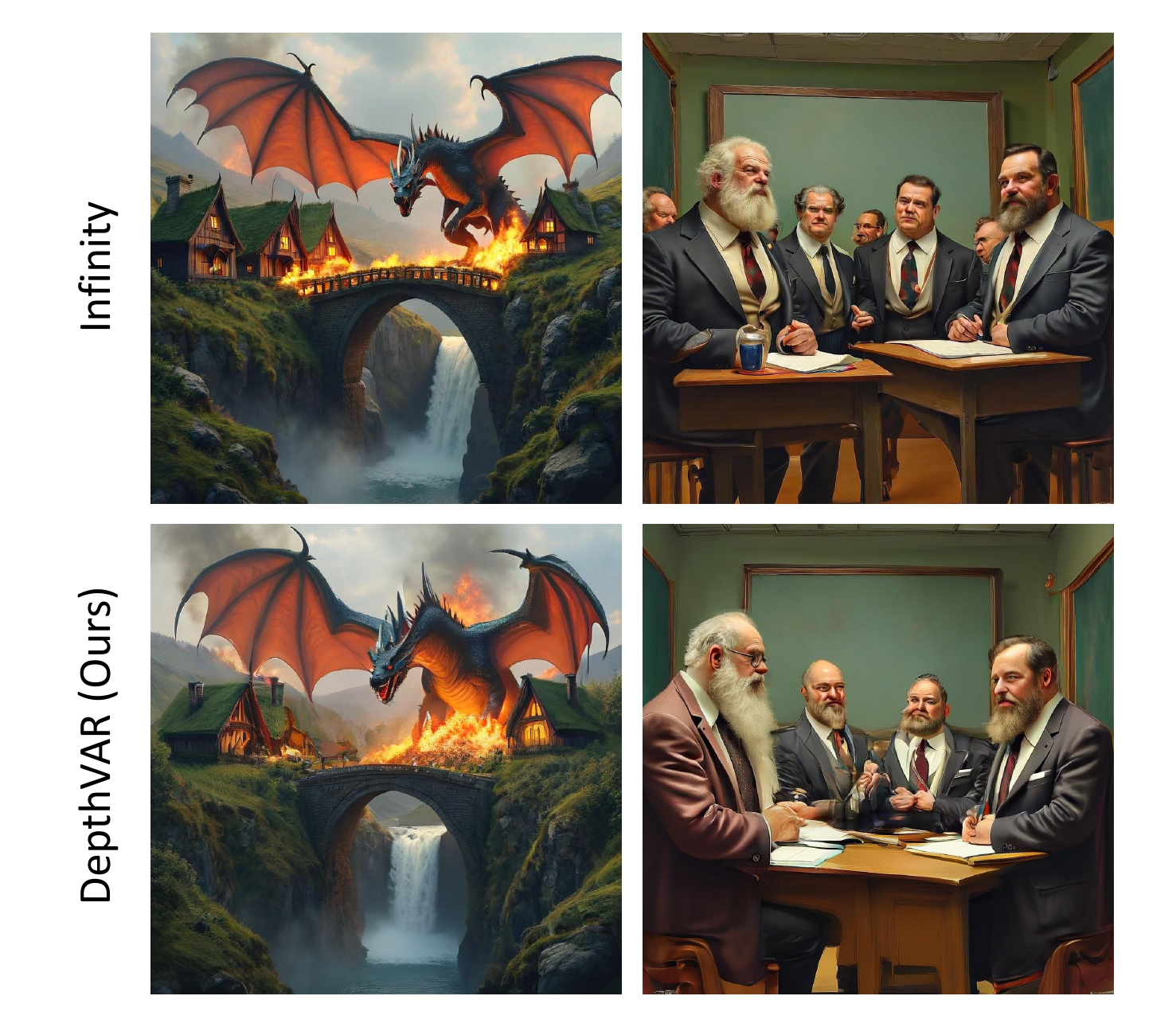}
    \caption{Finer Detail Loss.}
    \label{fig:failure_fg}
  \end{subfigure}
  
  \begin{subfigure}[t]{0.85\linewidth}
    \includegraphics[width=\linewidth, trim=38 0 10 0, clip]{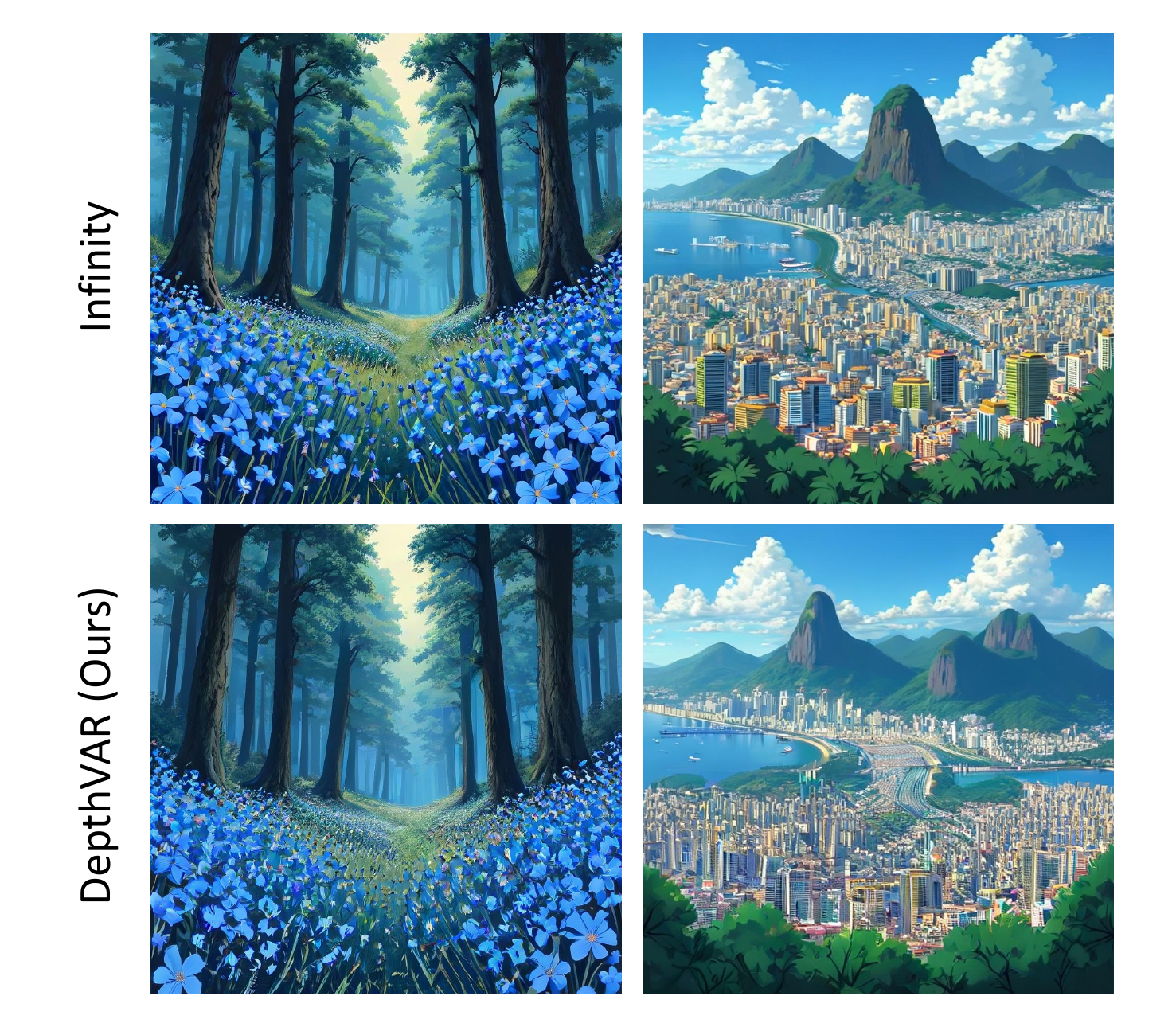}
    \caption{Dense Structure Handling.}
    \label{fig:failure_bg}
  \end{subfigure}
  \caption{Qualitative failure cases. (a) Loss of fine-grained details. (b) Difficulty with complex dense structures.}
  \label{fig:failure_case}
  \vspace{-1em}
\end{figure}

\begin{figure*}[ht]
  \centering
  \includegraphics[width=0.86\textwidth, trim=20 40 0 0, clip]{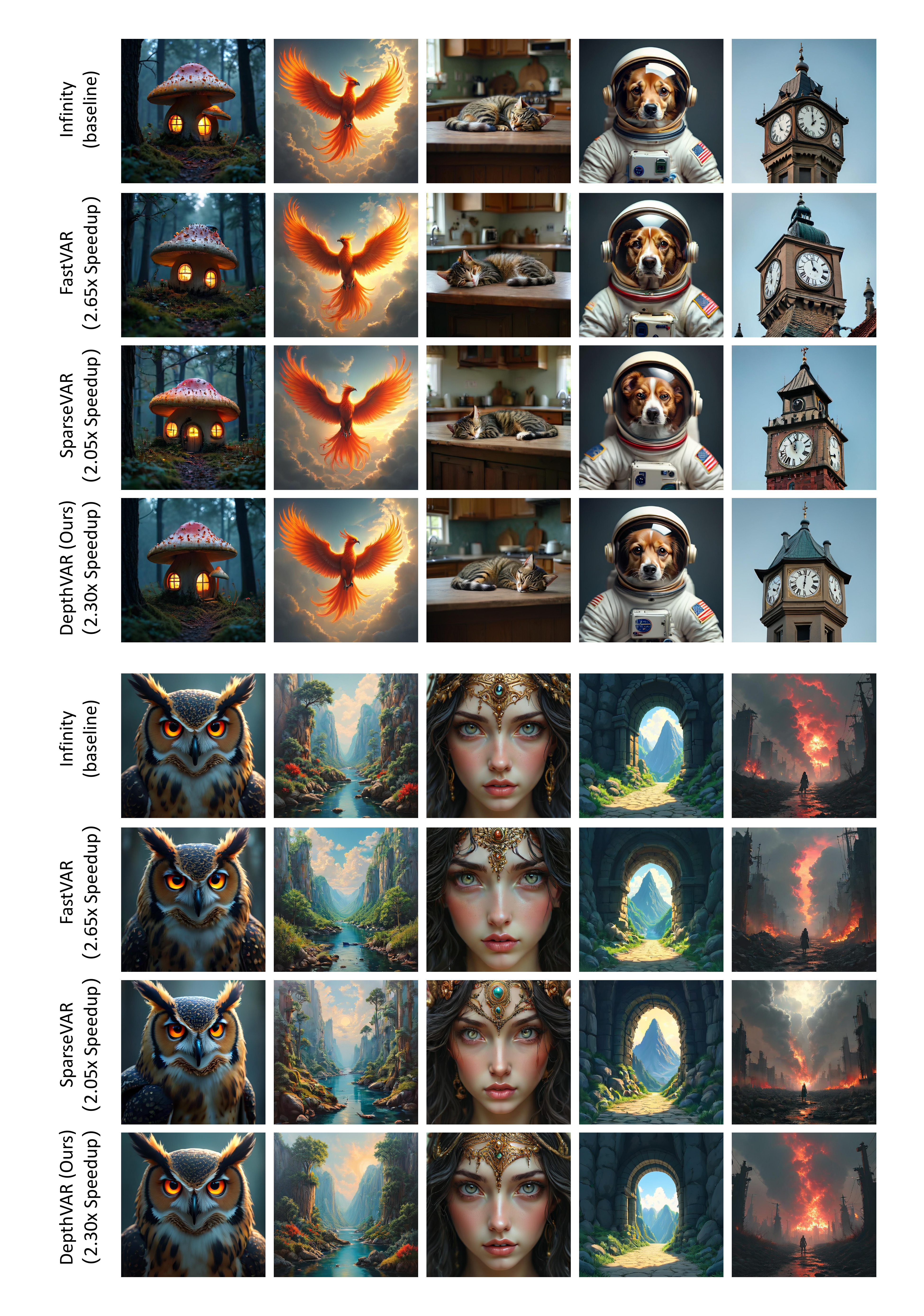}
  \caption{Additional qualitative visual comparisons from the HPSv2.1 benchmark. DepthVAR consistently preserves visual fidelity and semantic details, demonstrating a superior speed-quality trade-off.}
  \label{fig:more_res}
\end{figure*}

The qualitative visualizations in the main text (\cref{fig:res}) are derived from the ImageReward~\cite{xu2023imagereward} benchmark. We provide additional results from HPSv2.1~\cite{wu2023humanhps} in \cref{fig:more_res}. These examples further demonstrate that DepthVAR preserves high visual fidelity and rich detail, reinforcing its superior speed-quality trade-off compared to hard-pruning methods.
Despite its performance, DepthVAR has limitations. As shown in \cref{fig:failure_case}, it can struggle with fine-grained details and dense structures, likely because its fixed compute budget is insufficient for universally complex images. This limitation points toward future work in dynamic compute allocation, as discussed in our conclusion.

\end{document}